\documentclass{article}

\usepackage{arxiv}

\usepackage[utf8]{inputenc} 
\usepackage[T1]{fontenc}    
\usepackage{hyperref}       
\usepackage{url}            
\usepackage{booktabs}       
\usepackage{amsfonts}       
\usepackage{nicefrac}       
\usepackage{microtype}      
\usepackage{lipsum}
\usepackage{graphicx}
\usepackage{subcaption}
\usepackage{float}
\usepackage{multirow}
\usepackage{multicol}
\usepackage{tabularx}
\usepackage{amsmath}
\usepackage{fixltx2e}
\usepackage{pdfpages}
\graphicspath{ {./Fig/} }

\usepackage{color}
\definecolor{corr}{rgb}{1,0,0}

\title{MIcro-Surgical Anastomose Workflow recognition challenge report}

\author{
 Arnaud Huaulmé \\
  Univ Rennes\\
  INSERM, LTSI - UMR 1099\\
  F35000, Rennes, France\\
  \texttt{arnaud.huaulme@univ-rennes1.fr} 
   \And
 Duygu Sarikaya \\
  Gazi University\\
  Faculty of Engineering\\
  Department of Computer Engineering\\
  Ankara, Turkey
  \And
  Kévin Le Mut \\
  Univ Rennes\\
  INSERM, LTSI - UMR 1099\\
  F35000, Rennes, France
  \And
  Fabien Despinoy \\
  Univ Rennes\\
  INSERM, LTSI - UMR 1099\\
  F35000, Rennes, France
  \And
  Yonghao Long \\
  Department of Computer\\ Science \& Engineering,\\
  T Stone Robotics Institute,\\
  The Chinese University of Hong Kong
  \And
  Qi Dou\\
  Department of Computer\\ Science \& Engineering,\\
  T Stone Robotics Institute,\\
  The Chinese University of Hong Kong
  \And
  Chin-Boon Chng\\
  National University of Singapore\\ (NUS), Singapore, Singapore.\\
  Southern University of Science and \\Technology (SUSTech), Shenzhen, China.
  \And
  Wenjun Lin\\
  National University of Singapore\\ (NUS), Singapore, Singapore.\\
  Southern University of Science and \\Technology (SUSTech), Shenzhen, China.
  \And
  Satoshi Kondo\\
  Konica Minolta, Inc
   \And
  Laura Bravo-Sánchez\\
  Center for Research and \\Formation in Artificial Intelligence,\\
  Department of Biomedical Engineering,\\
  Universidad de los Andes, Bogotá, Colombia
  \And
  Pablo Arbeláez\\
  Center for Research and \\Formation in Artificial Intelligence,\\
  Department of Biomedical Engineering,\\
  Universidad de los Andes, Bogotá, Colombia
  \And
  Wolfgang Reiter\\
  Wintegral GmbH
  \And
  Manoru Mitsuishi\\
  Department of Mechanical Engineering\\
  the University of Tokyo\\
  Tokyo 113-8656, Japan
  \And
  Kanako Harada\\
  Department of Mechanical Engineering\\
  the University of Tokyo\\
  Tokyo 113-8656, Japan
  \And
  Pierre Jannin\\
  Univ Rennes\\
  INSERM, LTSI - UMR 1099\\
  F35000, Rennes, France\\
  \texttt{pierre.jannin@univ-rennes1.fr} 
}

\begin{document}
\maketitle
\begin{abstract}
Automatic surgical workflow recognition is an essential step in developing context-aware computer-assisted surgical systems. Video recordings of surgeries are becoming widely accessible, as the operational field view is captured during laparoscopic surgeries. Head and ceiling mounted cameras are also increasingly being used to record videos in open surgeries. This makes videos a common choice in surgical workflow recognition. Additional modalities, such as kinematic data captured during robot-assisted surgeries, could also increase the workflow recognition rate.

The “MIcro-Surgical Anastomose Workflow recognition on training sessions” (MISAW) challenge provided a data set of 27 sequences of micro-surgical anastomosis on artificial blood vessels. This data set was composed of videos, kinematics, and workflow annotations. The latter described the sequences at three different granularity levels: phase, step, and activity. The participants were given the option to use kinematic data and videos to develop workflow recognition models. Four tasks were proposed to the participants: three of them were related to the recognition of surgical workflow at three different granularity levels, while the last one addressed the recognition of all granularity levels in the same model. One ranking was made for each task. Additionally, to evaluate whether the recognition of several granularity levels could improve the recognition rate of each individual granularity, multi-granularity recognition models were also ranked with the uni-granularity ones. We used the average application-dependent balanced accuracy (AD-Accuracy) as the evaluation metric. This takes unbalanced classes into account and it is more clinically relevant than a frame-by-frame score.

Six teams, including a non-competing team, participated in at least one task. All models employed deep learning models, such as convolutional neural networks (CNN), recurrent neural networks (RNN), or a combination of both. The best models achieved more than 95\% AD-Accuracy for phase recognition, 80\% for step recognition, 60\% for activity recognition, and 75\% for all granularity levels. the RNN-based models outperformed the CNN-based ones as well as the dedicated modality models compared to the multi-granularity ones for phase and step recognition. For activity recognition, the multi-granularity models had better recognition rates.

For high levels of granularity (i.e., phases and steps), the best models had a recognition rate that may be sufficient for applications such as prediction of remaining surgical time or resource management. However, for activities, the recognition rate was still low for applications that can be employed clinically. The MISAW data set is publicly available to encourage further research in surgical workflow recognition. It can be found at \url{www.synapse.org/MISAW}

\end{abstract}

\keywords{Surgical Process Model \and Workflow recognition \and Multi-modality \and OR of the future}

\section{Introduction}

Computer-assisted surgical (CAS) systems should ideally make use of a complete and explicit understanding of surgical procedures. To achieve this, a surgical process model (SPM) can be used. A SPM is defined as a "simplified pattern of a surgical process that reflects a predefined subset of interest of the surgical process in a formal or semi-formal representation'' \cite{Jannin2001}. The SPM methodology is used for various applications, such as operating room optimization and management \cite{Sandberg2005,Bhatia2007}, learning and expertise assessment \cite{Huaulme2018,Forestier2018}, robotic assistance \cite{Ko2007}, decision support \cite{Quellec2015}, and quality supervision \cite{Huaulme2020}.

According to Lalys et al. \cite{Lalys2013}, a surgical procedure can be decomposed on several levels of granularity such,e.g., phases, steps, and activities. Phases are the decomposition of a surgical procedure into the main periods of intervention (e.g., resection). Each phase is broken down into multiple steps corresponding to a surgical objective (e.g., to resect the pouch of Douglas). A step is composed of several activities that describe the physical actions (namely action verbs,e.g., cut) performed on specific targets (e.g., the pouch of Douglas) by specific surgical instruments (e.g., a scalpel). This initial definition was improved at a lower granularity level to take into account information closed to kinematic data \cite{Despinoy2015}: surgemes and dexemes. A surgeme represents a surgical motion with explicit semantic meaning (e.g., grab), and a dexeme is a numerical representation of the sub-gestures necessary to perform a surgeme.

In early publications \cite{Sandberg2005,Bhatia2007,Huaulme2018,Forestier2018,Ko2007,Quellec2015,Huaulme2020,Despinoy2015}, SPMs were manually acquired by human observers. However, this solution has several drawbacks: It is costly concerning human resources, time-consuming, observer-dependent, and errors could be made. In \cite{Huaulme2019}, the authors noted that for the annotation of a peg transfer task, the mean duration to manually annotate one minute of video was around 13 minutes, and 65 annotation errors were counted for 60 annotations although the task was less susceptible to subjective interpretation than a surgical operation. To overcome these issues, \cite{Huaulme2019} proposed an automatic annotation method based on the information extracted from a virtual reality simulator. Even though this is a promising solution to limit human annotation, it requires information that could be complicated to obtain in surgical practice, such as the interactions between the instruments and anatomical structures. Other solutions are currently being studied to reduce the amount of manual annotation as transfer learning from simulated data to real data \cite{Zisimopoulos2017} or from a limited amount of annotated data \cite{DiPietro2019}.

Despite these innovative methods, automatic and online recognition of surgical workflows is mandatory to bring context-awareness CAS applications inside the operating room. Various machine learning and deep learning methods have been proposed to recognize different granularity levels such as phases \cite{Bhatia2007,Padoy2010,Twinanda2017} , steps \cite{Bouarfa2011,James2007} ,and activities \cite{Ko2007,Lalys2012}. According to the type of surgery, different modalities could be used for workflow recognition. For manual surgery, unless it is possible to add multiple sensors, workflow recognition is generally restricted to video-only modalities \cite{Bhatia2007,Bouarfa2011,Lalys2012}. In the case of robot-assisted surgery (RAS), kinematic information is easily available. It is expected that multi-modal data will lead to easier automatic recognition methods, as is the case for the combination of video and eye gaze information \cite{James2007} or the combination of video and kinematic information based on RAS data \cite{Zappella2013SurgicalData}. However, some methods based on RAS data sets propose video-only methods \cite{Sarikaya,Funke2019} or kinematic-only methods \cite{Despinoy2015,DiPietro2019}. 

The “MIcro-Surgical Anastomose Workflow recognition on training sessions” (MISAW) challenge provided a unique data set for online automatic recognition of multi-granularity surgical workflows using kinematic and stereoscopic video information on a micro-anastomosis training task. The participants were challenged to develop uni-granularity (with phases, steps, or activities) and/ or multi-granularity workflow recognition models.

\section{Methods: Reporting of Challenge Design}
This section describes the challenge design through an explanation of the organization, the mission, the data set, and the assessment method of the challenge.

\subsection{Challenge organization}
The MISAW challenge was a one-time event organized as part of EndoVis for MICCAI2020 online. It was organized by five people from three different institutions: Arnaud Huaulmé, Kévin Le Mut, and Pierre Jannin from the University of Rennes (France), Duygu Sarikaya from Gazi University (Turkey), and Kanako Harada from the University of Tokyo (Japan). The challenge was partially funded by the ImPACT Program of the Council for Science, Technology and Innovation, Cabinet Office, Government of Japan. All challenge information was made available to the participants through the Synapse platform: \url{www.synapse.org/MISAW} .

Participation in the challenge was subject to the following policies: Participants had to submit a fully automatic method using kinematic and/or video data. The data that could be used for the training were restricted to the data provided by the organizers and publicly available data sets, including pre-trained networks. The publicly available data sets only covered data that were available to everyone when the MISAW data set was released. The results of all participating teams were announced publicly on the challenge day. Challenge organizers and people from the organizing institutions could participate but were not eligible for the competition.

The participating teams had to provide the following elements: the method's outputs, a write-up, and a Docker image allowing the organizers to verify the outputs provided. Due to the COVID-19 crisis, a pre-recorded talk was also mandatory to limit technical issues during the challenge day (online event). All technical information (how to create a Docker image, the output format, etc.) was provided to the participants during the challenge on the challenge platform. The participants could submit multiple results and Docker images. However, only the last submission was officially counted to compute the challenge results. No leader-board or evaluation results were provided before the end of the challenge.

The challenge schedule was as follows: The training and the test data sets were released on June 1st and August 24th 2020 respectively. Submissions were accepted until September 23rd (23:59 PST). The results were announced October 4th during the online MICCAI2020. The complete data set was released with this paper at: \url{www.synapse.org/MISAW}

The organizers' evaluation scripts were publicly available on the challenge platform. Participating teams were encouraged (but not required) to provide their code as open access.

\subsection{Mission of the challenge}
The objective of the challenge was to automatically recognize the workflow of an anastomose performed during training sessions using video and kinematic data. The challenge was composed of four different tasks according to the granularity level recognized. Three of these tasks were uni-granularity surgical workflow recognition, i.e., the model had to recognize one of the three available granularity levels (phase, step, or activity). The last task was a multi-granularity surgical workflow recognition, i.e., recognition of the three granularities with the same model.

The challenge data were provided by a robotic system used to realize micro-surgical anastomosis on artificial blood vessels through a stereoscopic microscope. Such micro-surgical anastomosis is performed in neurosurgery and plastic surgery. The surgical robotic technologies developed for micro-surgical anastomosis can be applied to other robotic surgeries requiring dexterous manipulation on small targets. Automatic recognition of this task is an essential step to help the realization of this task or to increase robotic autonomy from manual to shared control or full automation \cite{Beer2014TowardInteraction}.

The final biomedical application was robotic micro-surgical suturing of the dura mater during endonasal brain tumor surgery. Both applications were similar in the use of a robotic system, the microscopic dimension of the targets, and the surgical gestures.

\subsection{Challenge data set}
The challenge data set was composed of 27 sequences of micro-surgical anastomosis on artificial blood vessels performed by 3 surgeons and 3 engineering students. It was divided into a training data set composed of 17 cases and a test data set composed of 10 cases. The splitting of the data set was done to have a similar ratio of expertise in each data set (Tableau ~\ref{tab:dataset}). A case was composed of kinematic data, a video, and workflow annotation. The latter was not provided to participants for the test cases. 

\begin{table}[H]
    \centering
    \begin{tabular}{|c||c|c|c|c||c|c|}
    \hline
         &\multicolumn{4}{c||}{Training cases}&\multicolumn{2}{c|}{Test cases}  \\\cline{2-7}
         Participant & Surgeon 2& Surgeon 3& Student 1& Student 2& Surgeon 1 & Student 3\\\hline
         nb case & 3 & 4& 6 & 4&4&6\\\hline
    \end{tabular}
    \caption{Training and test case splitting}
    \label{tab:dataset}
\end{table}

\subsubsection{Data acquisition}

The video and kinematic data were synchronously acquired at 30 Hz by a high-definition stereo-microscope (960x540 pixels) and a master-slave robotic platform \cite{Mitsuishi2013}, respectively, by the Department of Mechanical Engineering of the University of Tokyo. The kinematic data were recorded by encoders mounted on the two robotic arms. The kinematic data consisted of x, y, z, $\alpha, \beta, \gamma$. The homogeneous transformation matrices for each robotic instrument were calculated as in equations \ref{equ:H_right} and \ref{equ:H_left}. The kinematic files also contained information about the grip and the output grip voltage.
\begin{equation}
    H_{right} = T_x(x)T_y(y)T_z(z)R_x (\frac{1}{18} \pi)R_y (\alpha)R_x(\beta- \frac{5}{9}\pi)R_y(\gamma))
    \label{equ:H_right}
\end{equation}
\begin{equation}
    H_{left} = T_x(x)T_y(y)T_z(z)R_x (-\frac{1}{18} \pi)R_y (\alpha)R_x(\beta+ \frac{1}{18}\pi)R_y(\gamma))
     \label{equ:H_left}
\end{equation}

\begin{table}[b]
\centering
\begin{tabular}{|c|c|c|c|c|}
\hline
   \multirow{2}{*}{Phases}&\multirow{2}{*}{Steps}& \multicolumn{3}{c|}{Activities}\\\cline{3-5}
   &&Verb&Target&Instrument\\\hline
   Suturing     & Needle holding &Catch&Needle&Needle holder \\
   Knot Tying   & Suture making &Give slack&Wire &\\
                & Suture handling & Hold&Both artificial vessel&\\
                & 1° knot &Insert&Left artificial vessel& \\
                & 2° knot &Loosen completely& Right artificial vessel&\\
                & 3° knot &Loosen partially& Long wire strand&\\
                &&Make a loop &Short wire strand&\\
                &&Pass through&Wire loop&\\
                &&Position& Knot&\\
                &&Pull&& \\\hline
                                           
\end{tabular}
\caption{MISAW vocabulary. }
\label{tab:vocabulary}
\end{table}

The workflow annotation was acquired manually by two non-medical observers from the MediCis team of the LTSI Laboratory from the University of Rennes. The observers used the software “Surgery Workflow Toolbox [annotate]” provided by the IRT b<>com \cite{Garraud2014} to annotate the phases, steps, and activities (action verb, target, and instrument) of each robotic arm according to an annotation protocol. The vocabulary contained 2 phases, 6 steps, 10 action verbs, 9 targets, and 1 surgical instrument (Table \ref{tab:vocabulary}). The protocol described how to recognize each phase, step, and activity of each robotic arm by giving a definition, start and end point, and graphical illustration. For example, the step "suture making" was defined by "insert and pull the needle into artificial vessels." The start point was the "beginning of the needle insertion on one vessel," the stop point was "the needle completely pass through both vessels." This is illustrated in Figure \ref{fig:suture_making}. The complete annotation protocol is available in Supplementary Material C.

\begin{figure}[H]
    \centering
    \includegraphics[width=.5\linewidth]{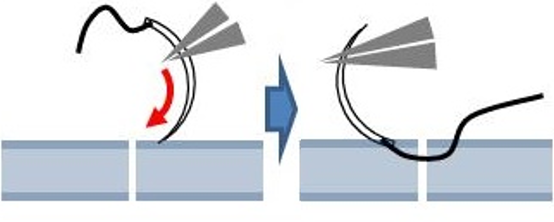}
    \caption{Representation of the beginning (left) and the end (right) of the "suture making" step.}
    \label{fig:suture_making}
\end{figure}

\begin{figure}[H]
    \centering
    \includegraphics[width=.8\linewidth]{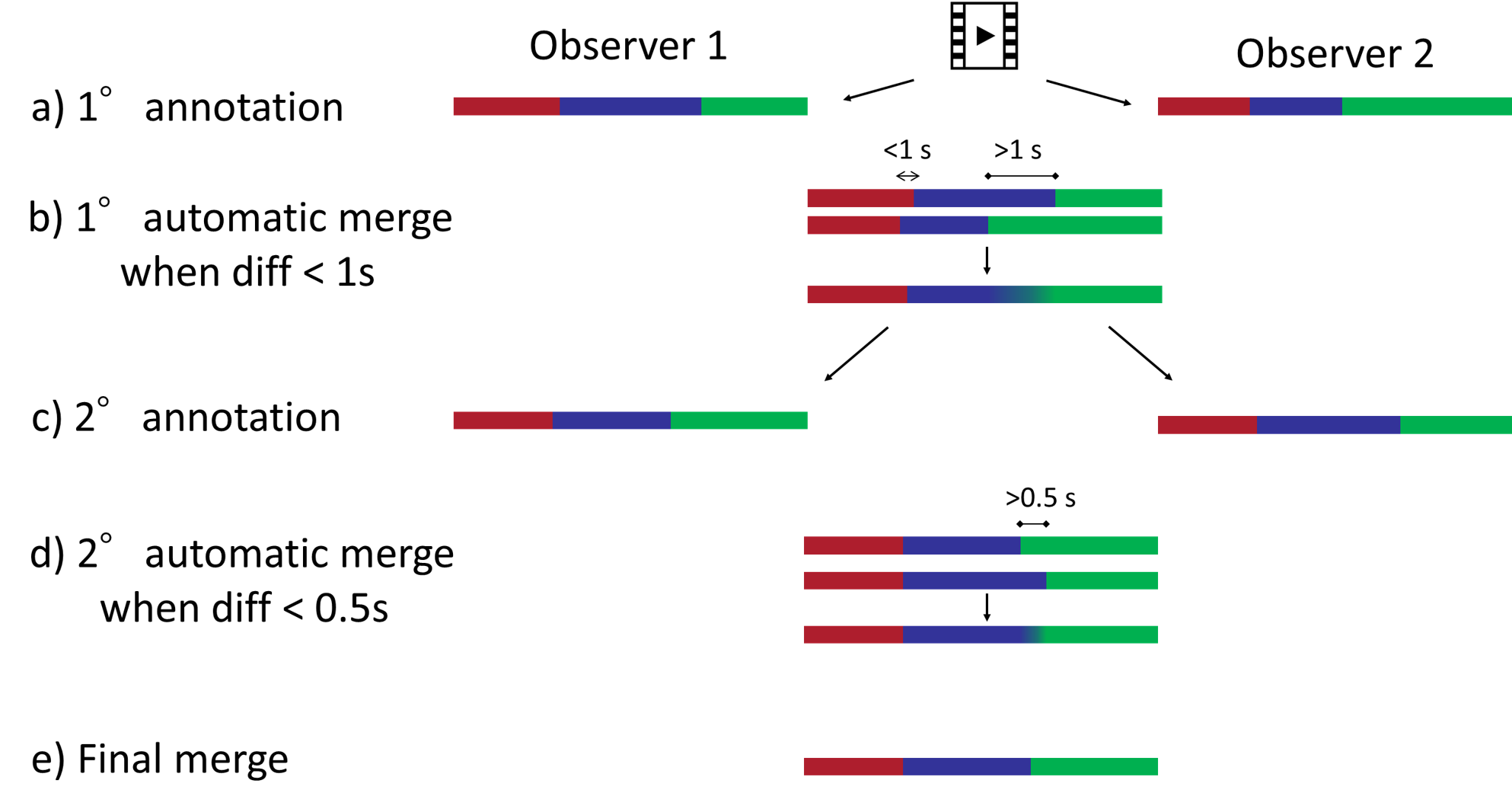}
    \caption{Harmonization protocol used to merge the annotations made by two observers.}
    \label{fig:workflowHarmonisation}
\end{figure}

Each case was annotated by both observers independently and harmonized by the following protocol (Figure \ref{fig:workflowHarmonisation}). An automatic merging was performed when the transition difference between both observers was less than one second (b in Figure \ref{fig:workflowHarmonisation}). Here, the transition between red and blue components was inferior to the threshold, so the automatic merging took the mean. The transition between the blue and the green components took longer than one second, so no decision was made. The merging sequence came back to each observer separately to refine uncertain transitions (c). A second automatic annotation was performed with a threshold of 0.5 seconds (d). Finally, all remaining uncertainties were harmonized by a consensus between both observers.

\subsubsection{Data pre-processing}
\begin{figure}[H]
    
    \begin{subfigure}[]{.5\textwidth}
        \centering
        \includegraphics[width=.8\linewidth]{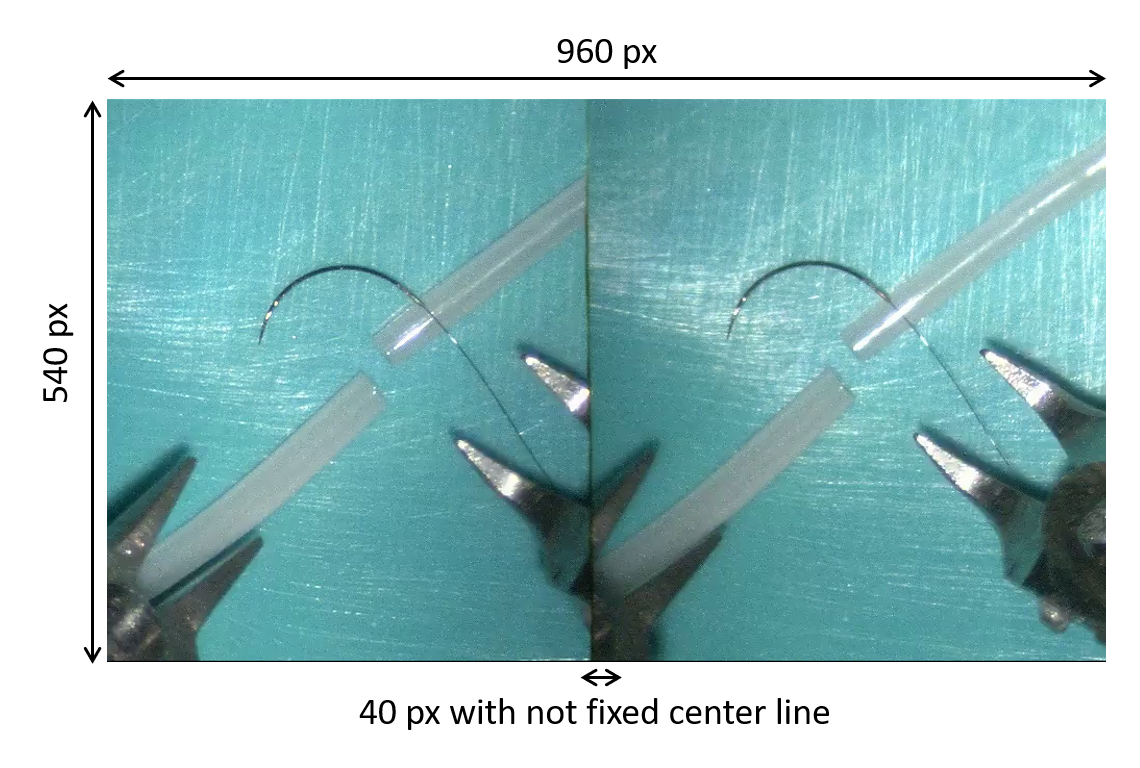}
        \caption{}
        \label{subfig:centerLineA}
    \end{subfigure}\hfill%
    \begin{subfigure}[]{.5\textwidth}
        \centering
        \includegraphics[width=.8\linewidth]{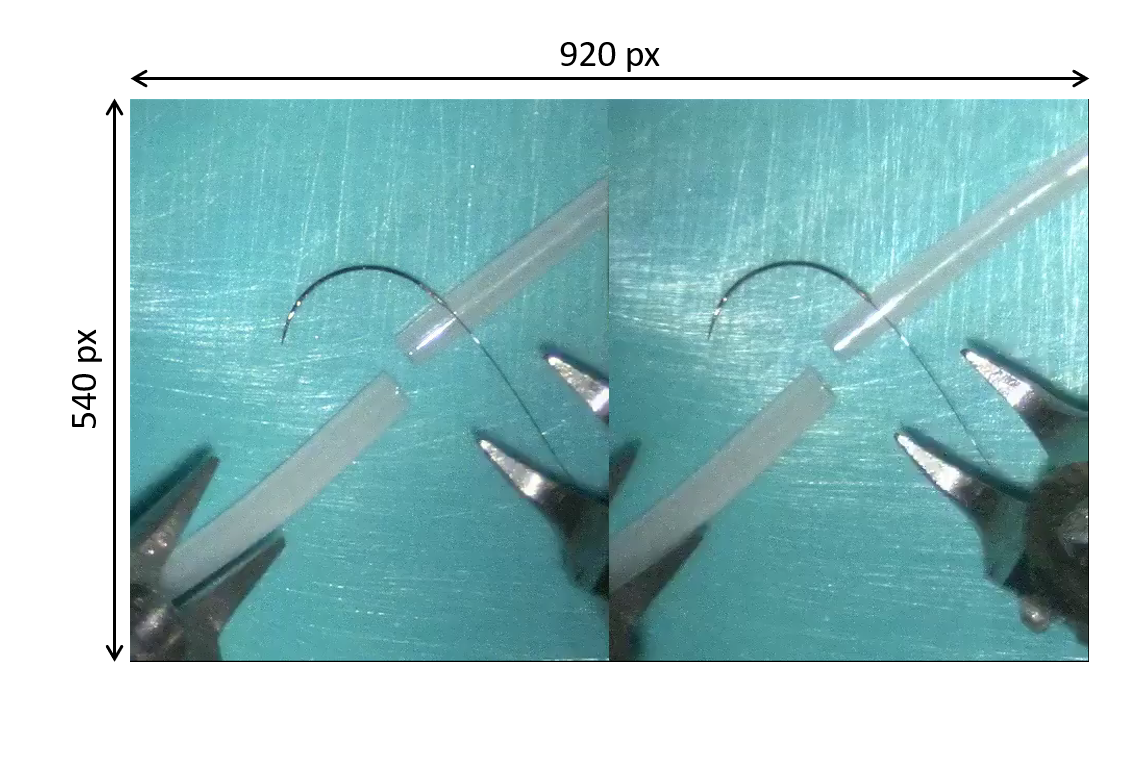}
        \caption{}
        \label{subfig:centerLineB}
    \end{subfigure}
    \caption{Comparison between the initial video (\ref{subfig:centerLineA}) and pre-processed video (\ref{subfig:centerLineB})}.
    \label{fig:centerLine}
\end{figure}

We pre-processed the videos and initial workflow annotations to have consistent and synchronized data for each case. In the videos, the boundary between the left image and the right image was not consistent (Figure \ref{fig:centerLine}, i.e., the position of the centerline was a little different within and between the trials. We removed 40 pixels from the center of the stereoscopic image to have two images of 460x540 pixels. The final video resolution was 920x540 pixels.

The software “Surgery Workflow Toolbox [annotated]” produced a description of sequences where each element was characterized by the beginning and the end in milliseconds. We modified it to provide a discrete sequence synchronized at 30Hz with the kinematic data. When no phase, step, or activity occurred, the term "Idle" was added. For each timestamp, we provided the following information: timestamp\_number, phase\_value, step\_value, verb\_Left\_Hand, target\_Left\_Hand, instrument\_Left\_Hand, verb\_Right\_Hand, target\_Right\_Hand, and instrument\_Right\_Hand.

\subsubsection{Sources of errors}
The main source of errors was the manual workflow annotation, which was observer-dependent. We limited these errors through the double annotations and the harmonization.

The second possible source of errors came from an acquisition issue. During acquisition, some timesteps were not acquired in either the video and kinematic information. This did not affect the synchronization of the data but could create activities not present in the procedural description. The impacted cases were 2-3, 4-2, 4-4, and 5-3.

Finally, due to some system problems during acquisition, the grip data were doubtful. If the system worked correctly, 0 meant "open" and -6 meant "close," but maybe values were under -6 in some trials.

These sources of errors were communicated to the participants with the training data set. The participants did not report any other issues.

\subsection{Assessment method}

\subsubsection{Metrics}
To assess the methods proposed by participants, we used a balanced version of the application-dependent scores \cite{Dergachyova2016} of the classic metric used in the workflow recognition: accuracy, precision, recall, and F1.

\begin{figure}[H]
    \centering
    \includegraphics[width=.8\linewidth]{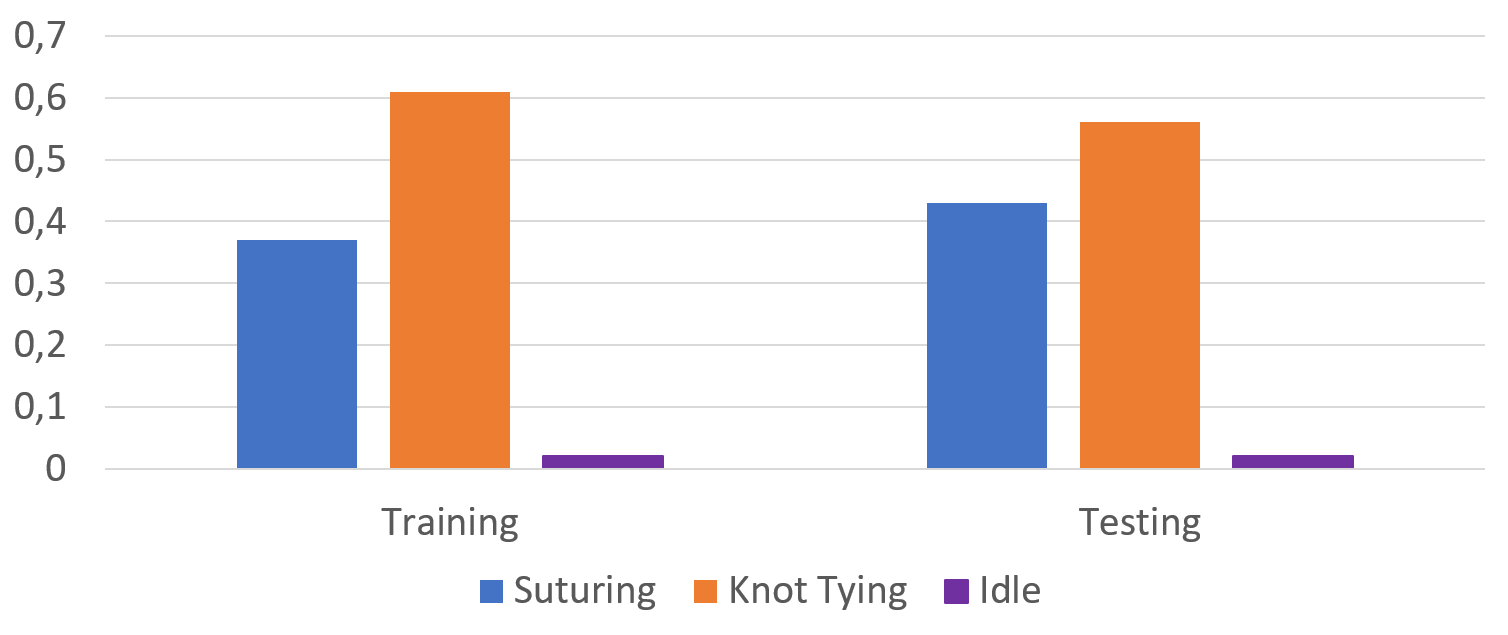}
    \caption{Phase distribution in training and testing data sets}
    \label{fig:phaseUnbalanced}
\end{figure}

Our data sets had a high class unbalance, for example, the phase "Idle" represented around 2\% of the frames in both data sets (Figure \ref{fig:phaseUnbalanced}). To give the same importance to each class, we decided to use balanced scores.

\begin{figure}[H]
    \centering
    \includegraphics[width=.8\linewidth]{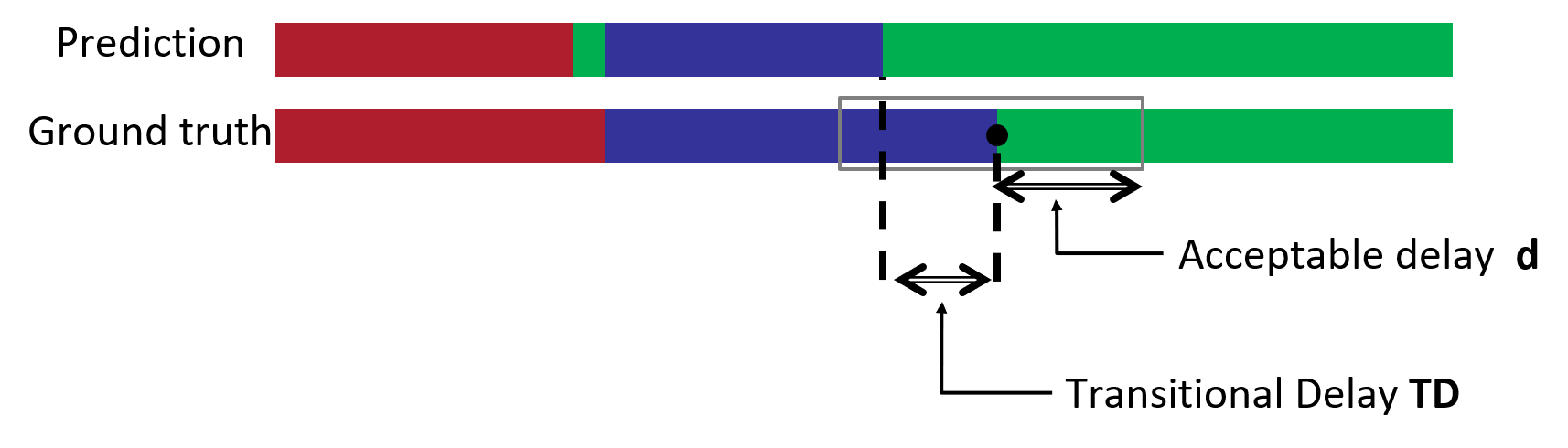}
    \caption{Definition of acceptable and transitional delay used to compute the average-dependent scores}
    \label{fig:ADScores}
\end{figure}

Generally, frame-by-frame scores were used. This type of score assumes that the ground truth is frame perfect. However, this is not possible with manual annotation. Moreover, a clinical application does not need to be 100\% accurate at a frame resolution. Application-dependent scores re-estimate classic scores using acceptable delay thresholds for a transitional window (Figure \ref{fig:ADScores}). When the transition on the predicted sequence occurs with a transition delay $TD$ inferior to an acceptable delay $d$ centered into the real transition, all frames are considered correct. Here, this was the case for the transition between the blue and green components. If the transition was different (case between red and green components in the prediction sequence) or outside this transition delay, no modification was done. We fixed the acceptable delay $d$ at 500 ms, which corresponded to half of the duration used for the first automatic merge (Figure \ref{fig:workflowHarmonisation}).

\subsubsection{Ranking method}\label{subsect:ranking}
We used a metric-based aggregation on the balanced application-dependent accuracy (AD-Accuracy) for the ranking. For each participant, we aggregated the metric values over all test cases and aggregated overall metrics to obtain a final score. We used a metric-based aggregation according to the conclusion of \cite{Maier-Hein2018}, who reported this type of aggregation as one of the most robust.

For the phase and step recognition (Tasks 1 and 2), the ranking score for algorithm $a_i$ was computed as follows:

\begin{equation}
    s_{uni}(a_i) = \frac{\sum_{t=0}^{T}balance\_accuracy\_case\_t}{T}
    \label{equ:s_uni}
\end{equation}

Activity recognition (Task 3) consisted of recognizing the action verb, target, and instrument of each robotic arm. The ranking metric was computed as follows, with each component (i.e.,  $s_{verb\_Left}(a_i)$) computed with Equation \ref{equ:s_uni}:
\begin{equation}
    \begin{split}
        s_{activity}(a_i) = \frac{1}{6} *( & s_{verb\_Left}(a_i)+ s_{Target\_Left}(a_i)+ s_{Instrument\_Left}(a_i)+ \\
        & s_{verb\_Right}(a_i)+ s_{Target\_Right}(a_i)+ s_{Instrument\_Right}(a_i))
    \end{split}
\end{equation}

For the multi-granularity recognition (Task 4) the ranking score was the mean of each uni-granularity score:
\begin{equation}
    s_{multi}(a_i) = \frac{s_{phase}(a_i)+s_{step}(a_i)+s_{activity}(a_i)}{3}
\end{equation}

All multi-granularity recognition models were also ranked in each uni-granularity task to highlight the differences between the models.

In the case of missing results, we considered results as good as a total random recognition. For example, for 3-class problem, the missing result would be set to 1/3, and for a 12-class problem, it would set to 1/12.

The ranking stability was assessed by testing different ranking methods. If, the ranking was not stable according to the method chosen, a tie between the different teams was pronounced.

The ranking computation and analysis were performed with the ChallengeR package provided by \cite{Wiesenfarth2021MethodsResults}.

\section{Results: Reporting of the Challenge Outcomes}
\subsection{Challenge submission}
At the end of September 2020, we counted 24 individual participants registered to the MISAW challenge and 325 downloads of the 9 files available (the Synapse platform did not give statistics by file). Five competing teams and one non-competing team completed their submissions for the challenge.

\subsection{Information on participating teams and corresponding methods}
In this section, we will present information on each team, the methods they used, and which tasks they participated in. The presentation is made in alphabetical order of the competing teams and not in consideration of their ranking.

\subsubsection{MedAIR}
The MedAIR team was composed of Yonghao Long and Qi Dou from the Department of Computer Science and Engineering at the Chinese University of Hong Kong. They participated in the phase and step recognition tasks.

The MedAIR team used both the video frames and the kinematic data of the left and right robotic arms, treating them separately because different arms may conduct different actions to jointly complete a task. 

They extracted high-level features from video frames using an 18-layer residual convolutional network \cite{He2016} followed by a fully connected layer and a ReLU non-linearity layer applied at the end, yielding a 128-dimension spatial feature vector. To learn the temporal information of the video data, they adopted a temporal convolution network (TCN) \cite{Lea2016},i.e., an encoder-decoder module, to further capture the information across frames, generating the representative spatial-temporal visual features. For the kinematic data, they first normalized the variables into [-1, +1], and then they used the TCN and long short-term memory (LSTM)  \cite{Dipietro2016} in parallel to learn and model the complex information of the left and right arms separately, yielding spatial-temporal motion features.

After acquiring the encoded high-level features from the video stream and kinematic data of the left and right arms, they used a graph convolutional network to further learn the joint knowledge among the multi-modal data. Considering that the visual information and left/right kinematic information contained fruitful interactions and relationships, they designed a graph convolutional network (GCN) with three node entities corresponding to the video, left kinematics, and right kinematics, with all three nodes connected to each other. Initialized with these three modalities, the node features of the GCN were updated by receiving the message from neighbor connected node features and then encoding stronger information in the newly generated node features. Then, the authors max pooled the features from the three nodes and forwarded them into a fully connected layer to get the prediction results of the workflow recognition. For more details, you can refer to \cite{Long2020RelationalSurgery}.

Two different approaches were employed to further enhance the temporal consistency of the workflow recognition. The authors filtered out the frames with low prediction probabilities using a median filter and leveraged the information of the preceding 600 frames with 30 fps. They also employed an online post-processing strategy (PKI) \cite{Jin2018} that leveraged the workflow of the phases and steps. For example, the steps followed a specific order: "Needle holding", "Suture making", "Suture handling", "1 knot", "2 knot", and "3 knot", and it were not likely to be reversed or shuffled.

\subsubsection{NUSControl Lab}
The NUSControl Lab team was composed of Chin Boon Chng\textsuperscript{1}, WenJun Lin\textsuperscript{1,2}, Jiaqi Zhang\textsuperscript{1}, Yaxin Hu\textsuperscript{1}, Yan Hu\textsuperscript{1}, Liu Jiang Jimmy\textsuperscript{2}, and Chee Kong Chui \textsuperscript{1}. The participants noted with "1" were from the National University of Singapore (NUS), Singapore, Singapore; the participants noted with "2" were from the Southern University of Science and Technology, Shenzhen, China. This team participated in the multi-granularity task. As described in the subsection "Ranking method" (\ref{subsect:ranking}), the model was also ranked in each uni-granularity task.

The NUSControl Lab team used both the video and kinematic data. They first extracted the features of the video frames using EfficientNet \cite{Tan2019}. Then, they employed an LSTM module to model the sequential dependencies of the video data. The authors hypothesized that the kinematic data were specifically related to the verbs and steps. With this motivation, they employed another LSTM module to model the sequential features of the left and right arm kinematic data, which was then concatenated and fed into a fully connected layer to predict the verbs (left and right) and the steps. Their network model was based upon the work of Jin et al. \cite{Jin2020}. 

The authors also employed a post-processing step that made used of the workflow observations to further improve the predictions. For example, if a knot is to be tied, a loop must first be made, followed by pulling the wire. Thus, the verb “make a loop” could be used to indicate when a new knot is being tied. Similarly, the verb “pull” could be used to indicate when the new knot has been completed. 
The authors proposed to mark the verb “make a loop” as a transition signal to the next knot and “pull” as a completion signal of this knot. If the model classified the current task to be “making a loop” and the phase turned to knot tying, the knot step was incremented. This knot step was identified to start from the previous “pull” prediction and continue until the next “pull” prediction.

\subsubsection{SK}
The SK team was composed of Satoshi Kondo from Konica Minolta, Inc. This team participated in the multi-granularity task. As described in the subsection "Ranking method" (\ref{subsect:ranking}), the model was also ranked in each uni-granularity task.

The SK team used the video data, kinematic data, and time information as the input for the model. The video frame features were extracted using a 50-layer ResNet \cite{He2016} pre-trained with the ImageNet data set, which led to a 2,048-dimension feature vector. While the team used only the left stereo video frame, the kinematic data features: x, y, z, $\alpha, \beta, \gamma$, and grip collected from the left and right arms were used, leaving the output voltage for the grip feature out. The kinematic data were normalized with the mean and standard deviation values for each dimension and then fed to two fully connected layers. 

The team also employed the frame number as a means of time information. The frame number was divided by 10,000. The feature vector of the input image, the feature vector of the kinematic data, and the frame number were concatenated, which led to a 2,063-dimension feature vector for a single frame. Then, the author performed multi-granularity recognition wherein the network learned the tasks,i.e., phase, step, and activity. For each activity, the verb, the target, and the tool for the left and right arms were learned, which resulted in a total of eight classes. The loss function was the summation of softmax cross-entropy for these eight classes, and the team employed a Lookahead optimizer \cite{Zhang2019}.

\subsubsection{UniandesBCV}
The UniandesBCV team was composed of Laura Bravo-Sánchez, Paola Ruiz Puentes, Natalia Valderrama, Isabela Hernández, Cristina González, and Pablo Arbeláez. All members were from the Center for Research and Formation in Artificial Intelligence and the Biomedical Computer Vision Group at the Universidad de los Andes, Colombia. This team participated in all proposed tasks.

The UniandesBCV team only used the video data and proposed a model that leveraged the implicit hierarchical information in the surgical workflow. The model presented by the authors was based on SlowFast \cite{Feichtenhofer2019}, a neural network that uses a slow and a fast pathway to model semantic and temporal information within videos. To accomplish this discrimination of information, each of the pathways analyzed the video at a different sampling rate. The slow pathway used a low frame rate with a large number of channels, while the fast pathway employed a high frame rate and only a fraction of the channels. To make a prediction based on the complete information (semantic and temporal), the fast pathway fused with the slow one using several lateral connections at different points of the network.

The authors first extracted the features of the video frames using ResNet-50 backbone \cite{He2016} and fed the features of 64-frame windows into a SlowFast model adapted for multi-task training that was pre-trained on the Kinetics data set \cite{Kay2017}. The authors explored different multi-task hierarchical groupings: The first model simultaneously predicted both phases and steps, the second model predicted activities, and the last model predicted all the components of the multi-granularity recognition. During training, the team also introduced an extra term to the loss function for optimizing the task added at each grouping and balanced the relevance of each task by associating each of the loss’s terms to a weight. The authors reported that merging all the components of the multi-granularity recognition tasks improved the learning ability of the model and obtained more accurate predictions.

\subsubsection{Wr0112358}
Team wr0112358 was composed of Wolfgang Reiter from Wintegral GmbH. This team participated in all proposed tasks.

Team wr0112358 only used the video data, reporting that the kinematic data did not significantly contribute to the performance of the model. The team also explored different architectures, including ResNet50 \cite{He2016} and multi-stream Siamese networks with temporal pooling \cite{Chung2017}, but reported that due to the high imbalance and the relatively small size of the data set, the complex architectures resulted in overfitting. The author also ruled out using an LSTM approach for the same reason. 

The team decided to employ a multi-task convolutional neural network \cite{Twinanda2017} and extracted the features of the video frames using a DenseNet121 CNN with data augmentation and regularization, which reduced overfitting. The author enhanced this architecture with task-wise early stopping \cite{Zhang2014} and also reported that using either of the stereo video frames resulted in a similar performance.

\subsubsection{IMPACT}
The IMPACT team was the non-competing team due to the presence of challenge organizers on it. This team was composed of Arnaud Huaulmé and Fabien Despinoy both from Rennes University, INSERM, LTSI - UMR 1099 and Duygu Sarikaya from Gazi University, Faculty of Engineering, Department of Computer Engineering. The team participated in all proposed tasks.

The IMPACT team used both the right video frames and kinematic data and proposed a multi-modal architecture. The authors applied a pre-processing step to the input data. While the right video images were rescaled from 460x540 to 224x224 and the pixel values were normalized by subtracting the mean of each channel over the training set and scale between [0,1], the authors applied a z-normalization to center the kinematic data to 0 with a standard deviation equal to 1. To make the training step faster, the data were downsampled to 5Hz.

Then, each input modality was processed into a dedicated network branch to leverage the different types of data. While the video frames were passed to a VGG19 network \cite{Simonyan2014Two-streamVideos}, the kinematic data were passed to an adapted ResNet network \cite{IsmailFawaz2018EvaluatingNetworks}. The last convolutional layer of each modality branch was finally concatenated into a common branch before being split again into separated workflow branches containing their own activation layers (1 for the phase and step recognition, 6 for the activity recognition, and 8 for the multi-granularity recognition).

The VGG19 network was initialized with the weights of a pre-trained model on the ImageNet data set. Since the MISAW data set was acquired in phantom surgical settings, the IMPACT team retrained only the last two layers to refine the network for this task. Regarding the kinematic branch, the network was trained "from scratch" without any previous weight configuration. In the end, the training was achieved using an Adam optimizer and a starting learning rate of 0.0001.

\subsection{Workflow recognition results}
Even if the participants submitted the method outputs for each test case, all of the following results were computed on organizer hardware via the provided Docker images. We did not detected any fraud attempts in the results provided by the participants.

This section only presents the results used for the ranking. Detailed results by sequence and task of each participating team are available in Supplementary Material B.

\subsubsection {Task 1: Phase recognition}
Phase recognition is a three-class task. We received 4 uni-granularity and 5 multi-granularity models for this task; the latter were identified with the addition of "\_multi" at the end of the team name.

\begin{figure}[H]
    \centering
    \includegraphics[width=\linewidth]{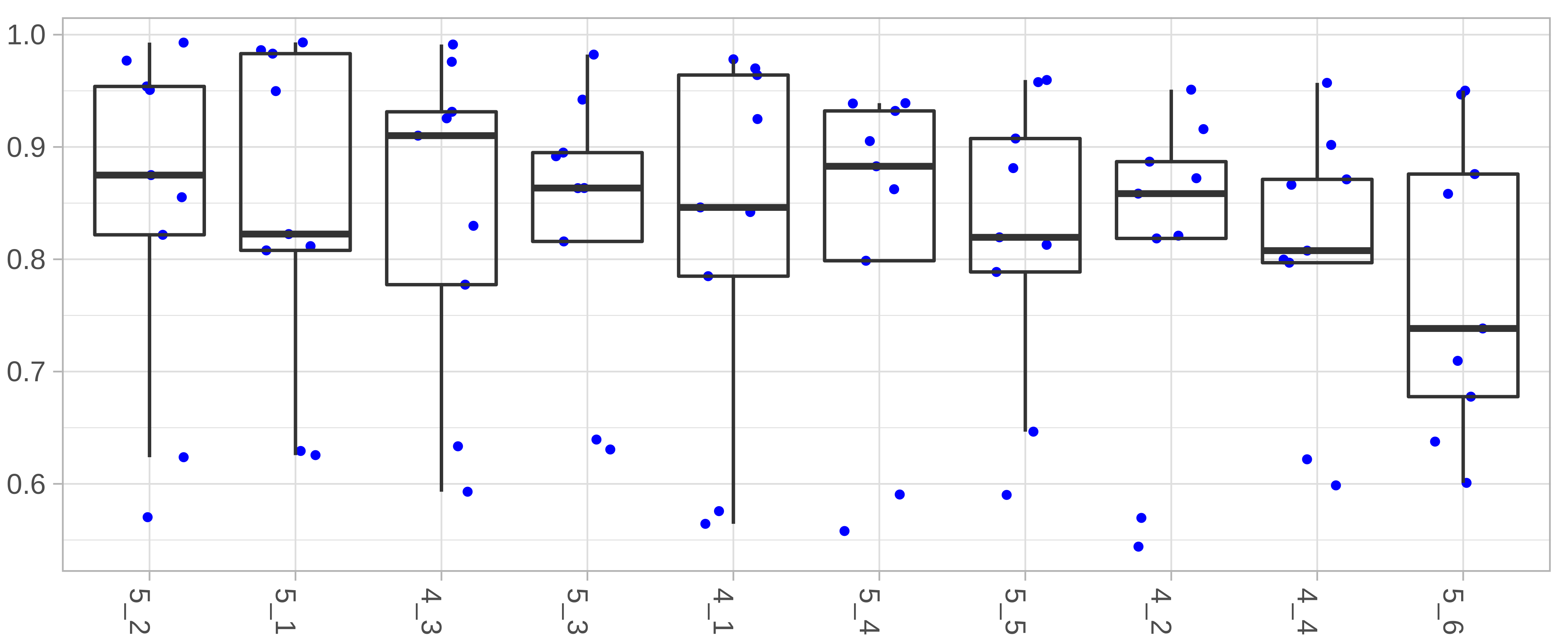}
    \caption{Phase recognition AD-Accuracy for each sequence. Each dot represents the AD-Accuracy for one model.}
    \label{fig:Phase_Results_seq}
\end{figure}

Figure \ref{fig:Phase_Results_seq} presents the results of all algorithms for each test sequence. The average AD-Accuracy by sequence was between 77.7\% and 84.7\%, which demonstrated that the recognition difficulty was similar between the sequences, except for sequence 5\_6. However, we noticed that for all the test cases, 2 models had an AD-Accuracy lower than 65\%.

\begin{figure}[H]
    \centering
    \includegraphics[width=\linewidth]{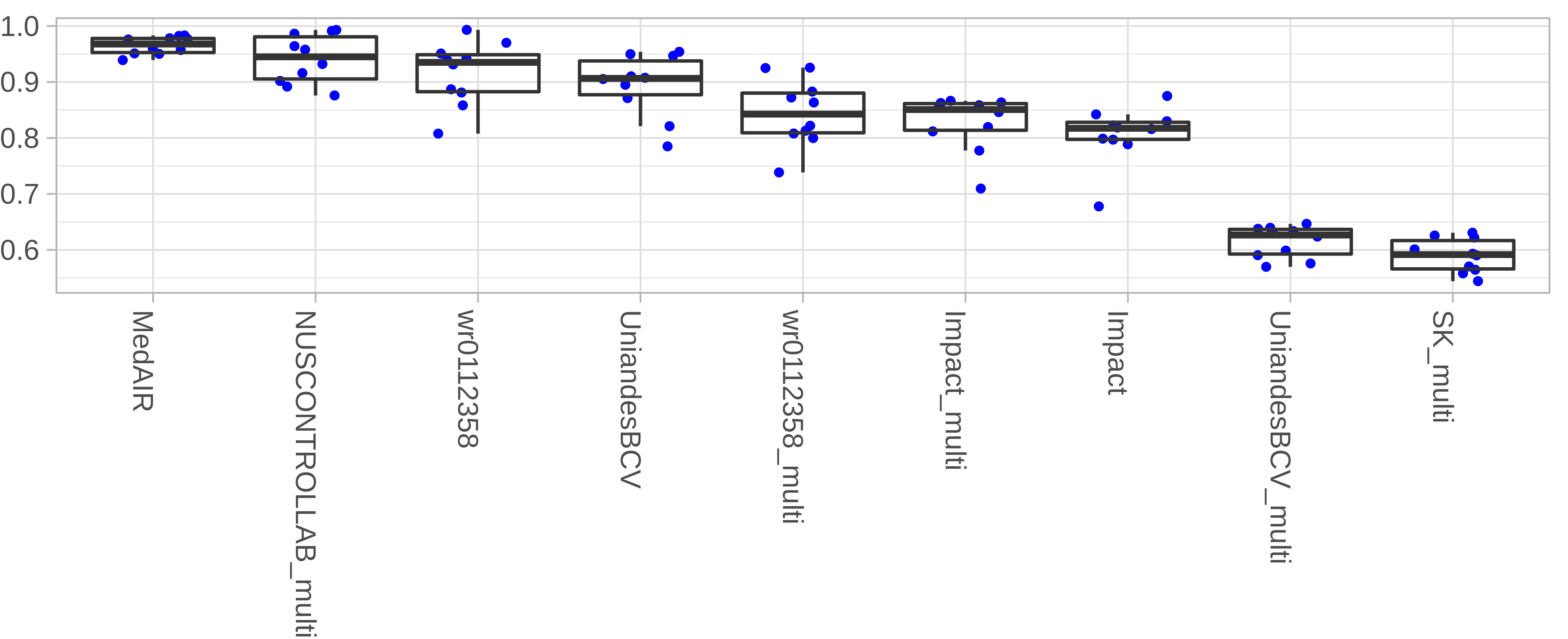}
    \caption{Average phase recognition AD-Accuracy for each model. Each dot represents the AD-Accuracy for one sequence.}
    \label{fig:Phase_Results}
\end{figure}

Figure \ref{fig:Phase_Results} presents the average AD-Accuracy for each model.The  MedAIR team got an average AD-Accuracy of 96.53\%. The multi-granularity models of the UniandesBCV and SK teams presented results lower than 65\%. Overall, only the uni-granularity model of IMPACT had an outlier lower than 70\%, while the average AD-Accuracy was 80.66\%. Is it also interesting to note that the multi-granularity model of this team was slightly better than the uni-granularity one.

\begin{figure}[H]
    \centering
    \includegraphics[width=\linewidth]{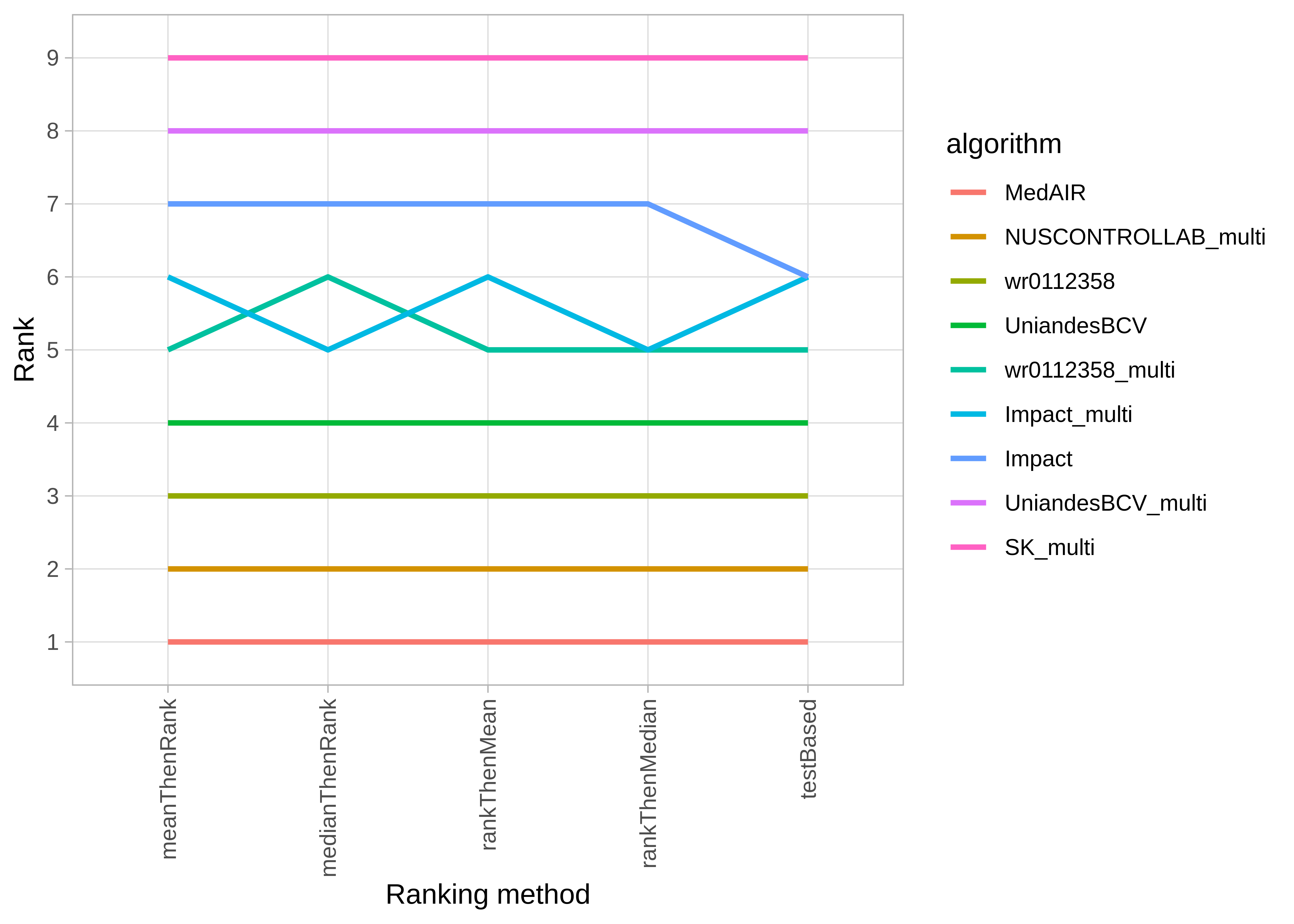}
    \caption{Phase recognition ranking stability through different ranking methods.}
    \label{fig:Phase_Ranking}
\end{figure}

Figure \ref{fig:Phase_Ranking} presents the different rankings according to the method chosen. For the phase recognition, the choice of method did not influence the ranking, except for the multi-granularity models of teams IMPACT and wr0112358, which swapped the fifth and sixth places.

\subsubsection {Task 2: Step recognition}
Step recognition is a 7-class task. We received 4 uni-granularity and 5 multi-granularity models for this task; the latter were identified with the addition of "\_multi" at the end of the team name.

\begin{figure}[H]
    \centering
    \includegraphics[width=\linewidth]{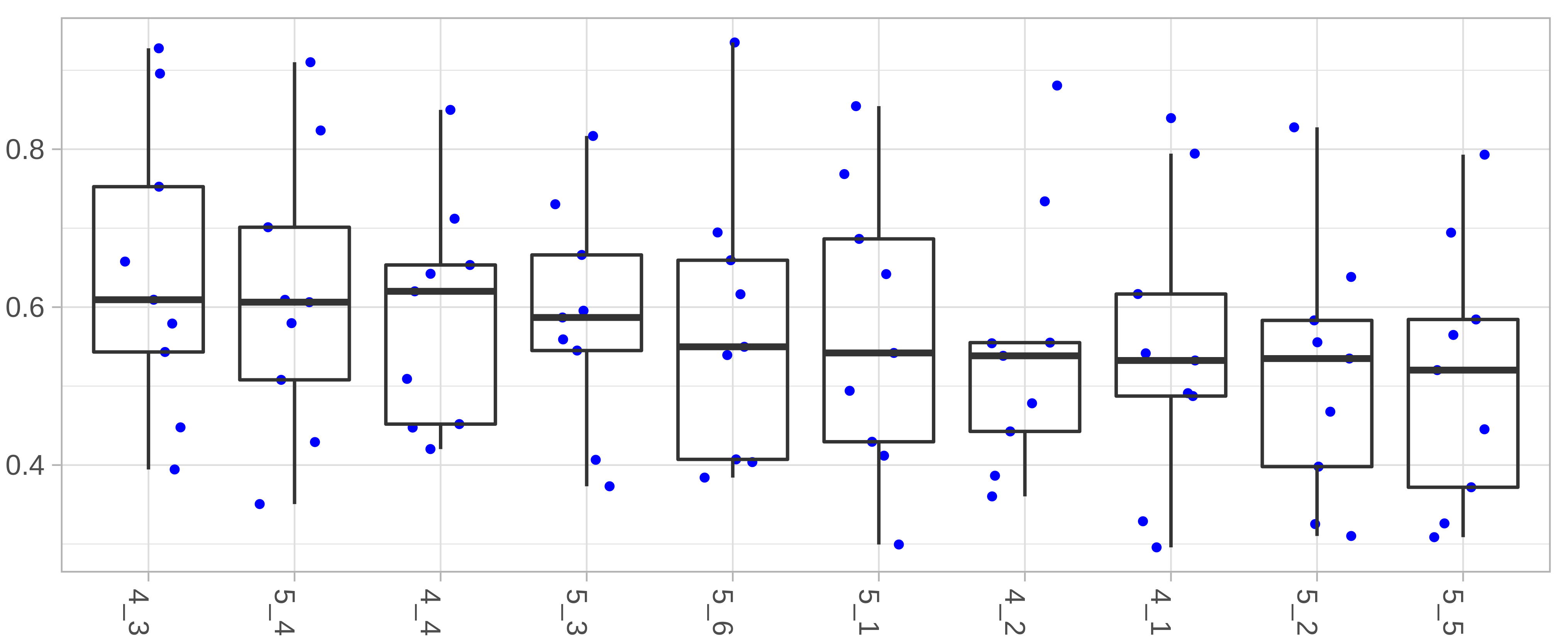}
    \caption{Step recognition AD-Accuracy for each sequence. Each dot represents the AD-Accuracy for one model.}
    \label{fig:Step_Results_Seq}
\end{figure}

The average AD-Accuracy by sequence was between 51.2\% and 64.4\% (Figure \ref{fig:Step_Results_Seq}). Contrary to the phase recognition, there was no sequence with a significantly lower score. We noticed that for all sequences, at least one model outperformed the others.

\begin{figure}[H]
    \centering
    \includegraphics[width=\linewidth]{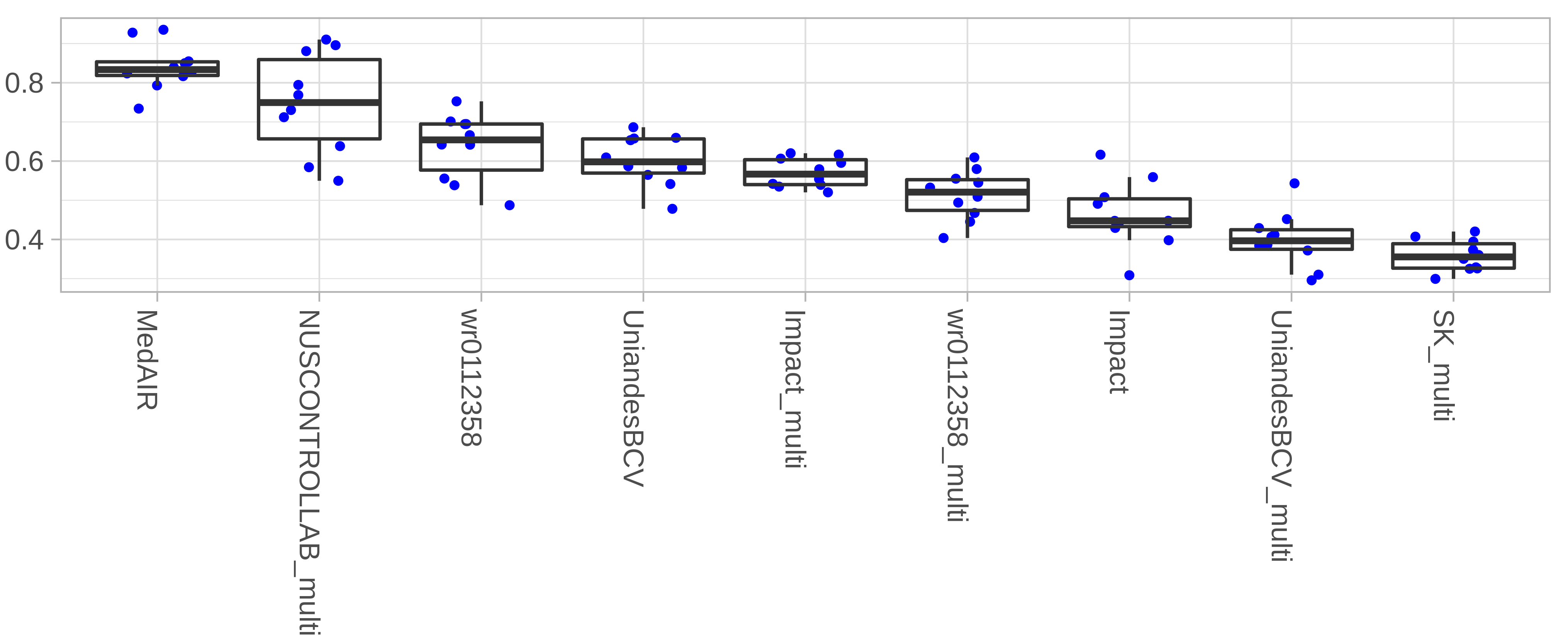}
    \caption{Average step recognition AD-Accuracy for each model. Each dot represents the AD-Accuracy for one sequence.}
    \label{fig:Step_Results}
\end{figure}

In figure \ref{fig:Step_Results}, we could identify this team as MedAIR, which obtained an average AD-Accuracy of 84.02\%. Three models had results lower than 50\%: the uni-granularity model of the IMPACT team and the multi-granularity models of the UniandesBCV and SK teams. Only the multi-granularity model of NUSControl Lab had disparate results according to the sequences.

\begin{figure}[H]
    \centering
    \includegraphics[width=\linewidth]{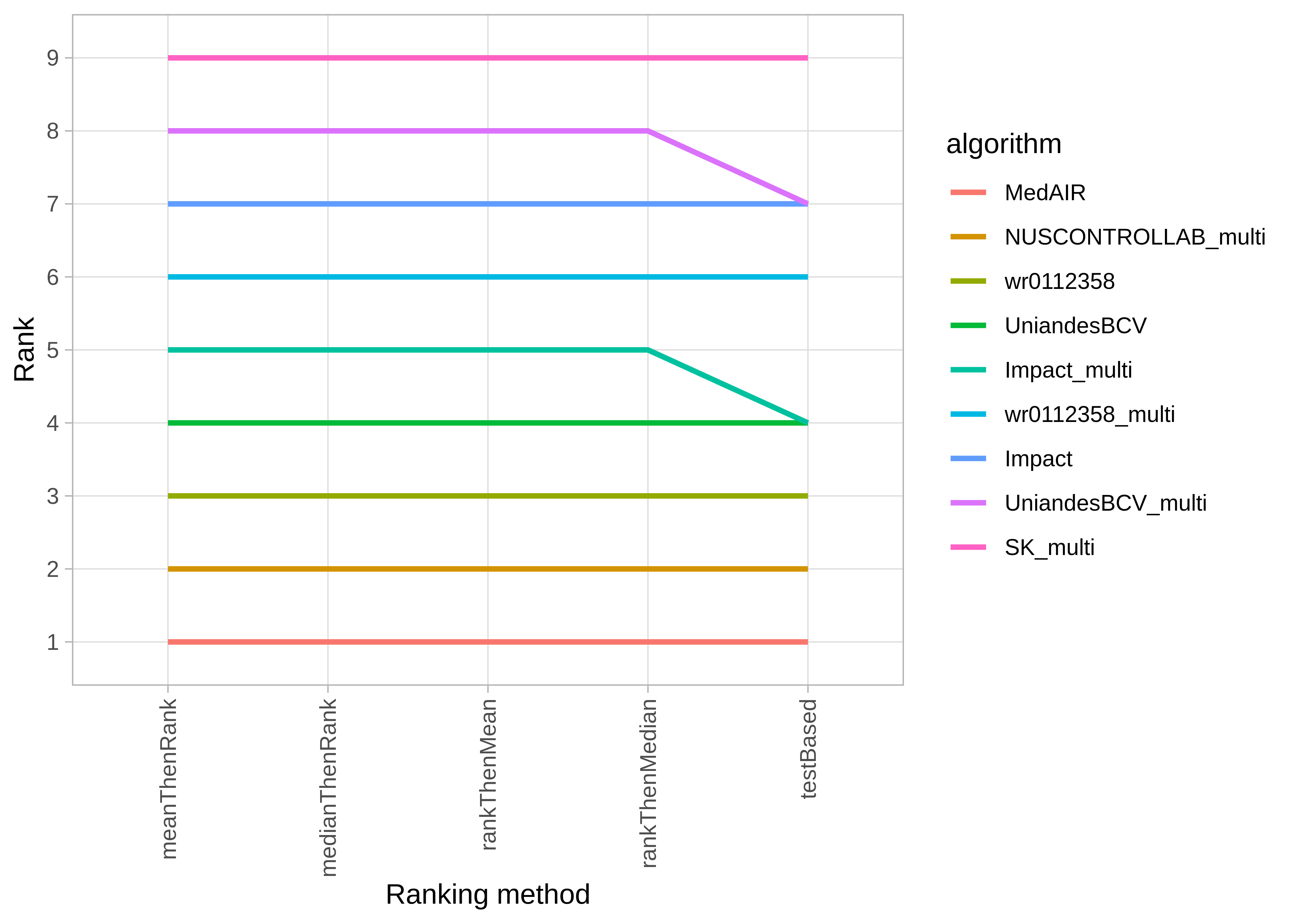}
    \caption{Step recognition ranking stability through different ranking methods.}
    \label{fig:Step_Ranking}
\end{figure}

The ranking method chosen did not impact the final rank (Figure \ref{fig:Step_Ranking}).

\subsubsection {Task 3: Activity recognition}
The activity recognition consisted of recognizing 6 components, i.e., the verb, target, and instrument for the left and right arms. Each component was an 11-, 10-, and 2-class problem, respectively. We received 3 uni-granularity and 5 multi-granularity models for this task; the latter were identified with the addition of "\_multi" at the end of the team name.

\begin{figure}[H]
    \centering
    \includegraphics[width=\linewidth]{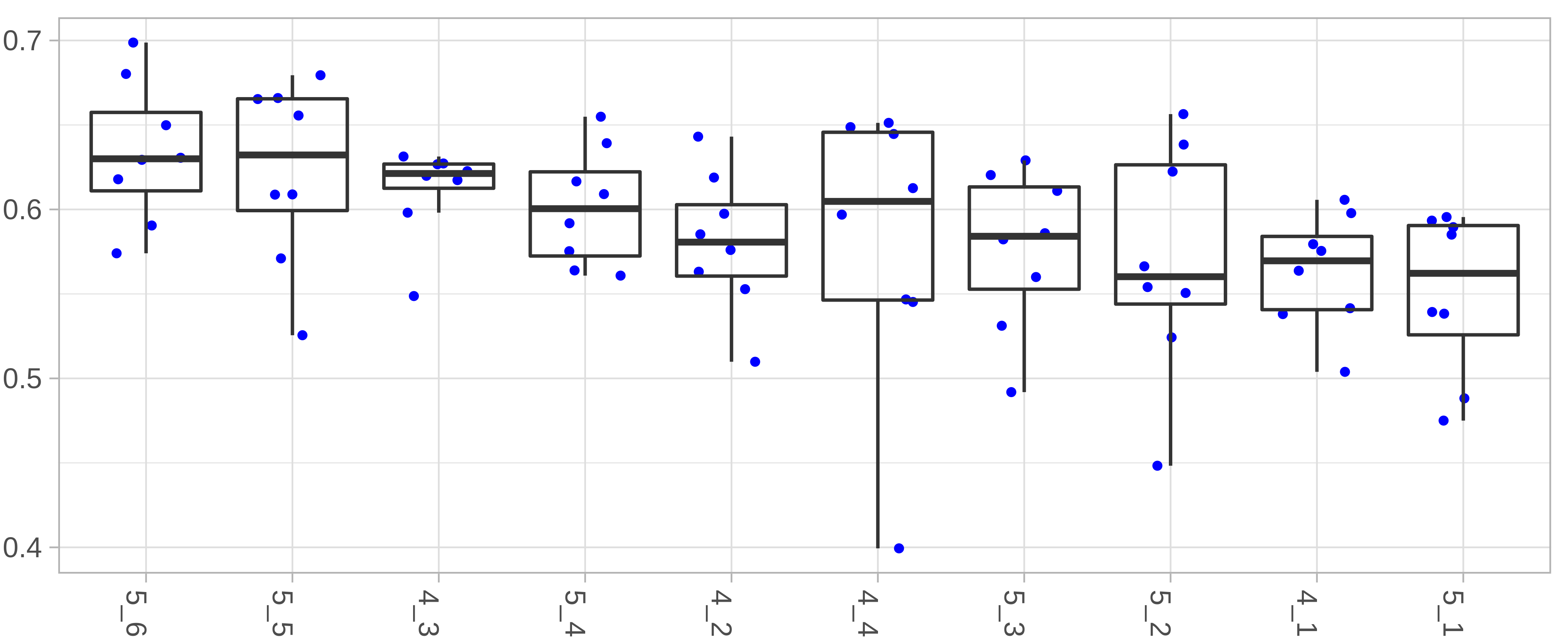}
    \caption{Activity recognition AD-Accuracy for each sequence. Each dot represents the AD-Accuracy for one model.}
    \label{fig:Activity_Results_Seq}
\end{figure}

The average AD-Accuracy score by sequence was between 55.1\% and 63.4\% (Figure \ref{fig:Activity_Results_Seq}). As for the step recognition, all sequences had similar results. However, for session 4\_4, one model had an AD-Accuracy lower than 40\%.

\begin{figure}[H]
    \centering
    \includegraphics[width=\linewidth]{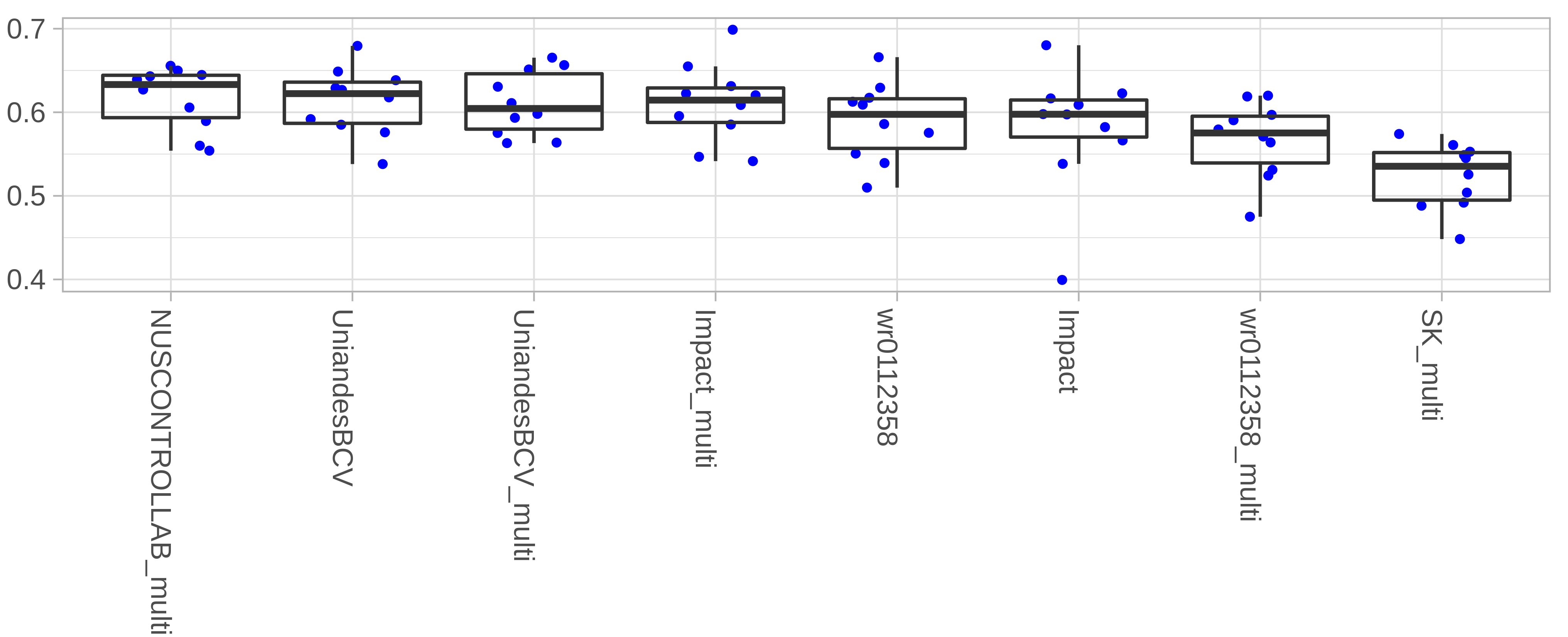}
    \caption{Average activity recognition AD-Accuracy for each model. Each dot represents the AD-Accuracy for one sequence.}
    \label{fig:Activity_Results}
\end{figure}

The average AD-Accuracy by model was between 52.4\% and 61.6\% \ref{fig:Activity_Results}. Four models, three of which were multi-granularity ones, had results over 60\%.

\begin{figure}[H]
    \centering
    \includegraphics[width=\linewidth]{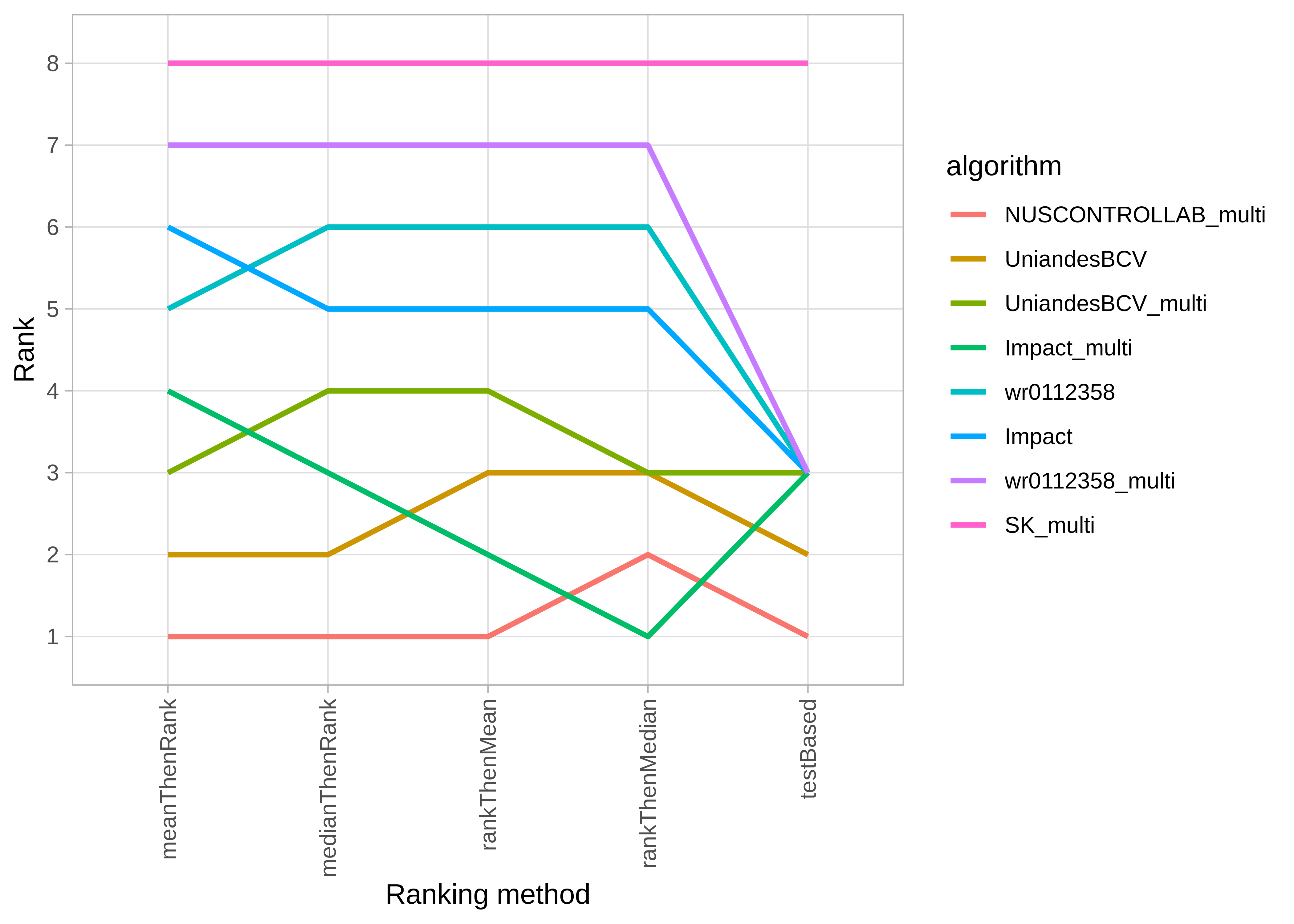}
    \caption{Phase recognition ranking stability through different ranking methods.}
    \label{fig:Activity_Ranking}
\end{figure}

According to the ranking method (Figure \ref{fig:Activity_Ranking}), the ranking was always different for the top four models. For this task, we defined a tie between the NUSControl Lab and UniandesBCV teams (IMPACT was a non-competitive team).

\subsubsection {Task 4: Multi-granularity recognition}
Task 4 consisted of recognizing the phase (a 3-class problem), the steps (a 7-class problem), and the verb, target, and instrument for the left and right arms (an 11-, 10-, and 2- class problem, respectively) on a unique model . Of the 6 teams, 5 proposed a model for this task.

\begin{figure}[H]
    \centering
    \includegraphics[width=\linewidth]{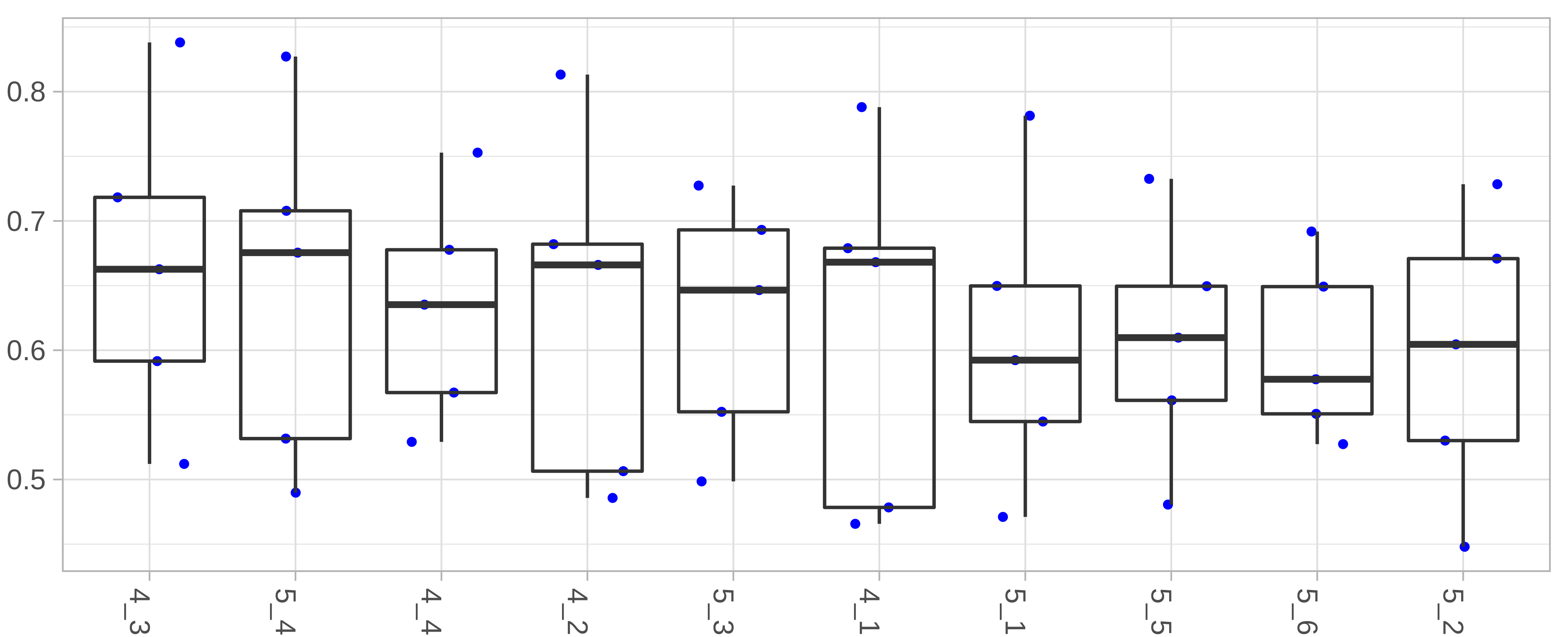}
    \caption{Multi-granularity recognition AD-Accuracy for each sequence. Each dot represents the AD-Accuracy for one model.}
    \label{fig:Multi_Results_Seq}
\end{figure}

The average AD-Accuracy score by sequences was between 59.6\% and 66.4\% (Figure \ref{fig:Multi_Results_Seq}). Surprisingly, these results were slightly better than those for the activity recognition although this task also demanded recognition of phases and steps.

\begin{figure}[H]
    \centering
    \includegraphics[width=\linewidth]{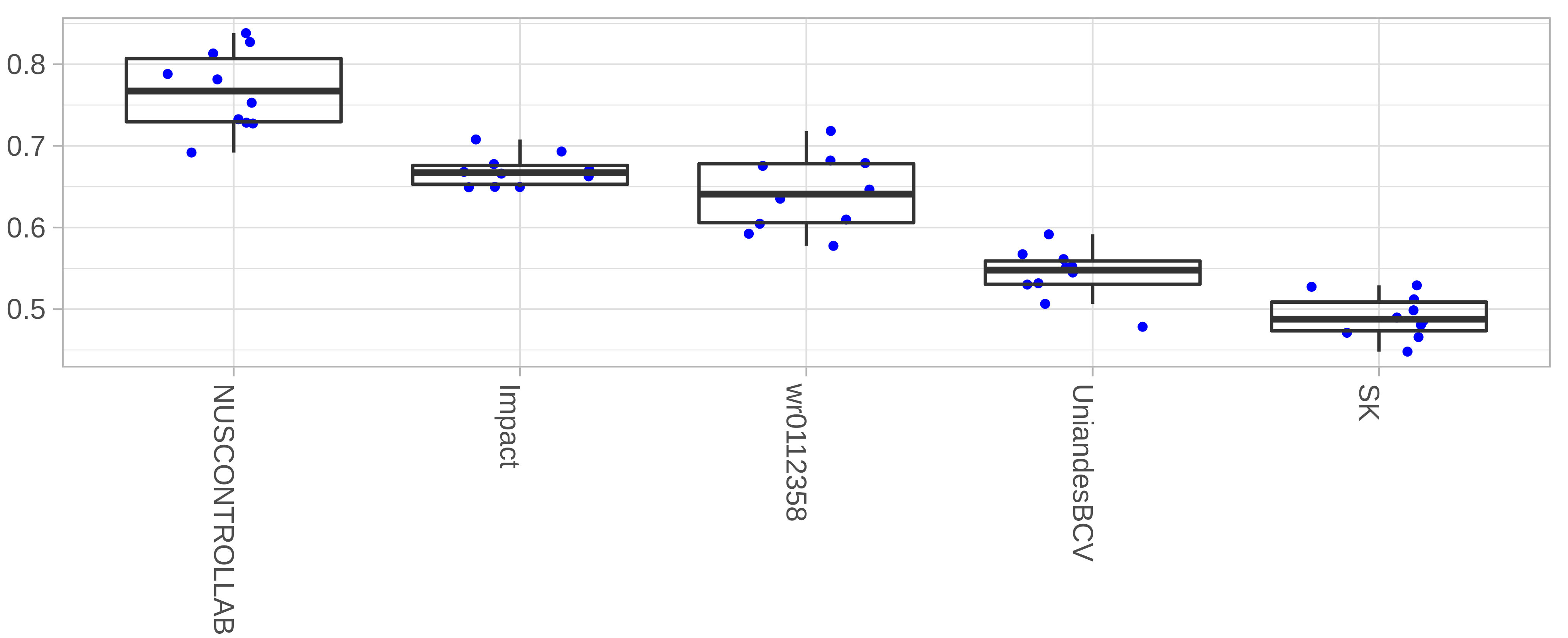}
    \caption{Average multi-granularity recognition AD-Accuracy for each model. Each dot represents the AD-Accuracy for one sequence.}
    \label{fig:Multi_Results}
\end{figure}

\begin{figure}[H]
    \centering
    \includegraphics[width=\linewidth]{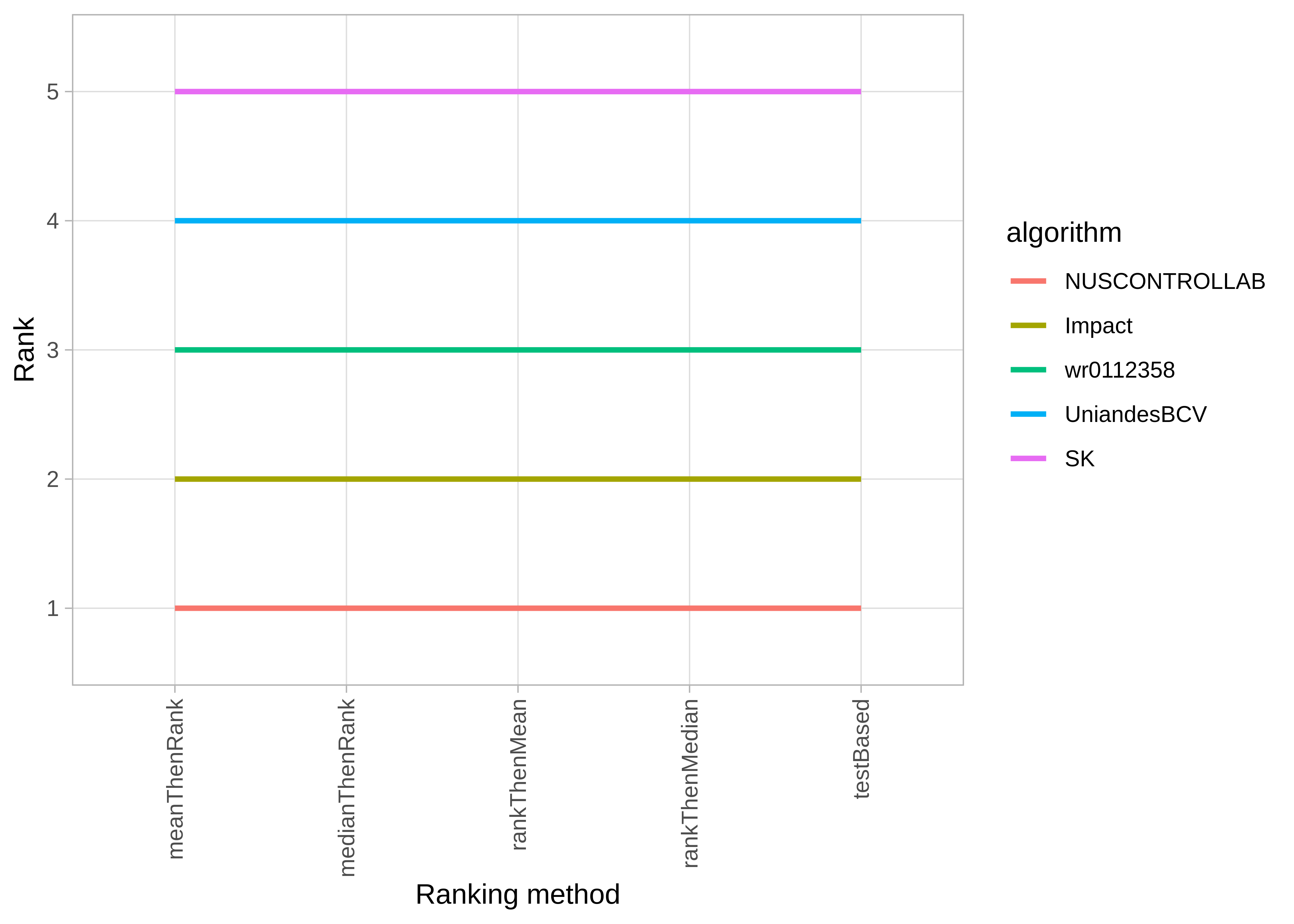}
    \caption{Multi-granularity recognition ranking stability through different ranking methods.}
    \label{fig:Multi_Ranking}
\end{figure}

The average AD-Accuracy by model was between 49.1\% and 76.8\% \ref{fig:Multi_Results}. The model of NUSControl Lab outperformed the models of the other teams, with a recognition rate 12 points higher than the second competing team (IMPACT had a better result than wr0112358 team but was not competing). The team ranking was not impacted by the ranking method (Figure \ref{fig:Multi_Ranking}).

\section{Discussion}
Surgical workflow recognition is an important challenge in providing automatic context-aware computer-assisted surgical systems. However, as demonstrated by the different models proposed in this challenge, there remains a lot of room for improvement. For a high level of granularity (phases and steps), the best models have a recognition rate that may be sufficient for applications such as prediction of remaining surgical time or resource management. However, for activities, the recognition rates are still insufficient to propose clinical applications.

For all tasks, the decrease between the sequence with the best recognition rate and the one with the lowest was linear. The difference between the best and the worst recognition rate was 7 points for phase recognition, 13.2 for step recognition, 8.3 for activity recognition, and 6.8 for multi-granularity recognition. Only sequence 5\_6 for phase recognition presented a recognition gap of 3 points with the penultimate sequence (Figure \ref{fig:Phase_Results_seq}). After a review of this sequence, the major difference was a high representation of the "idle" phase (7.13\% for sequence 5\_6 compared to 2.49$\pm$ 1.22\% for the other test cases) to the detriment of the "suturing" phase (36.12\% compared to 45.14$\pm$10.36\%) for a similar total duration (79s vs. 99s $\pm$ 52s). However, this cannot be the only reason for this low recognition rate. In the future, it could be interesting to study the explainability of the different networks.

For the image modality, all teams proposed a model based on convolutional neural networks (CNNs) such as  ResNet, and VGG, two of them also combined a recurrent neural network (RNN) as LSTM. For the kinematic modality, two teams used CNN, one used RNN, and another used a combination of both. The teams wr0112358 and UniandesBCV did not use this modality. According to the results, the use of RNNs seemed to be more relevant than that of CNNs only. However, both teams that used them also performed post-processing to improve the recognition rate, so it was difficult to evaluate the role of the RNNs and post-processing.

For the phases and step recognition tasks, the multi-granularity models had globally worse performances than the uni-granularity models, even for the teams who proposed both models on the same architecture. The only exception was for IMPACT, but the results were quite similar between the team's models. For the activity recognition, it was the opposite: 3 of the 4 top models were multi-granularity ones. Two reasons could explain this fact. First, the activity and multi-granularity models had to recognize multiple components at the same time (6 and 8 components, respectively), whereas the other models only recognized 1 component. Second, the majority of activities could only appear on a specific phase or step. For example, the activity consisting of inserting a needle on the right artificial vessel with a needle holder (noted <insert, right artificial vessel, needle holder>) could only appear on the phase "suturing" and, specifically, on the step "suture making". So, the multi-granularity models could learn these relations to improve their performances for activity recognition.

One of the most surprising results of the challenge was the similar recognition rate between the video-based models and the multi-modality-based models (using both videos and kinematics). Team wr0112358 reported that the kinematic data did not significantly contribute to the performance of their model. This was confirmed by the ranking of this team (top 3 for the phase and step recognition tasks with a dedicated model and top 3 for the multi-granularity task). The UniandesBCV team also used only the video modality, and also had good ranking, especially for activity recognition, with a tie for first place between both models proposed. However, it is impossible to know whether these results come from the models used by the participants or from the lack of information provided by the kinematic data. A more robust and systematic study would clarify this by the understanding of the models and the contribution of each modality.

The first main limitation was the unbalanced distribution of cases by expertise level (11 performed by experts, 16 by engineering students) due to the different number of cases by participants (between 3 and 6 cases). We split the data set to have a similar distribution between the training and test data sets to limit the impact of this unbalanced distribution.

The second main limitation was the release of the video and kinematic data of the test cases during the challenge. This choice was dictated by the organizers' lack of knowledge of Docker images and the lack of hardware available when the challenge was proposed to EndoVis and MICCAI. So, we wanted to be able to use the results provided by the participants if necessary. Finally, all results were computed on the organizers' hardware via Docker images. With the test cases release, we first asked unnecessary works to teams; the time spent running the results could have been dedicated to the improvement of the methods. Moreover, this early release could have allowed the participants to make their own manual annotations and use them for the training. Even if these annotations were different than those by the organizer, it opened a breach for undetectable fraud.

In addition to confirming the superiority of RNNs compared to CNNs with same post-processing method and studying the impact of each modality, future work could explore more complex networks such as hierarchical models. Indeed, the granularity description is hierarchic (a step belongs to a phase; some activities only appear on specific steps), so this type of model could improve the recognition. Enlarging the data set with more sessions, more modalities, and more sources of data (other systems, virtual reality simulators, etc.) is also being considered for a second version of the MISAW challenge.

\section*{Acknowledgements}
\noindent
This work was partially by ImPACT Program of Council for Science, Technology and Innovation, Cabinet Office, Government of Japan. 

\noindent
Authors thanks the IRT b\textless \textgreater com for the provision of the software ``Surgery Workflow Toolbox [annotate]'' , used for this work.

\bibliographystyle{unsrt}
\bibliography{MISAW}

\section*{Supplementary material}
\subsection*{A. Authors' contributions}
A. Huaulmé was the challenge coordinator, the primary contact with the participant teams, and a member of the IMPACT team. He made the workflow annotations, collected, computed, and analyzed the results, and wrote the paper. D. Sarikaya was a member of the challenge organizers and the IMPACT team and wrote the paper. K. Le Mut was a member of the challenge organizers and made the workflow annotations. K. Harada was a member of the challenge organizers and supervised, with M. Mitsuishi, the video and kinematic data acquisition. P. Jannin was the challenge supervisor. F. Despinoy was a member of the IMPACT team. Y. Long and Q. Dou were members of the MedAIR team. C.-B. Chng and W. Lin were members of the NUSControlLab team. S. Kondo was a member of the SK team. L. Bravo-Sánchez and P. Arbeláez was members of the UniandesBCV team. W. Reiter was a member of Wr0112358 team. All co-authors participated in the proofreading of the parts concerning their work.

\subsection*{B. Detailed results for each team}
\label{supMat:detailed_results}
The results for each team by sequence and task are presented on the following pages. Eight balanced scores were computed, the frame-by-frame and application-dependent version of the accuracy, precision, recall, and F1.

\subsubsection*{B.1 MedAIR}

\begin{table}[H]
    \centering
    \begin{tabular}{ | c | c |  c | c | c || c  | c | c | c |}
     \hline
          \multirow{2}{*}{Sequence}&\multicolumn{4}{c||}{Frame-by-Frame}&\multicolumn{4}{c|}{Application-Dependent}  \\\cline{2-9}
          &Accuracy&Precision&recall&F1&Accuracy&Precision&recall&F1\\\hline
          4\_1&97.81&97.65&96.83&97.08&97.81&97.65&96.83&97.08\\\hline
          4\_2&95.10&95.93&93.81&94.53&95.10&95.93&93.81&94.53\\\hline
          4\_3&97.59&98.00&96.71&97.14&97.59&98.00&96.71&97.14\\\hline
          4\_4&95.71&94.88&94.03&94.20&95.71&94.88&94.03&94.20\\\hline
          5\_1&98.31&97.99&97.61&97.72&98.31&97.99&97.61&97.72\\\hline
          5\_2&97.69&97.37&96.58&96.84&97.69&97.37&96.58&96.84\\\hline
          5\_3&98.23&98.13&97.63&97.78&98.23&98.13&97.63&97.78\\\hline
          5\_4&93.90&93.56&91.36&92.02&93.90&93.56&91.36&92.02\\\hline
          5\_5&95.96&94.64&94.01&94.10&95.96&94.64&94.01&94.10\\\hline
          5\_6&95.02&96.79&96.48&96.50&95.02&96.79&96.48&96.50\\\hline
          Mean &96.53&96.49&95.50&95.79&96.53&96.49&95.50&95.79\\\hline
     \end{tabular}
      \caption{Results of team MedAIR for phase recognition}
     \label{supMat:MedAIR_Phase}
\end{table}

\begin{table}[H]
    \centering
    \begin{tabular}{ | c | c |  c | c | c || c  | c | c | c |}
     \hline
          \multirow{2}{*}{Sequence}&\multicolumn{4}{c||}{Frame-by-Frame}&\multicolumn{4}{c|}{Application-Dependent}  \\\cline{2-9}
          &Accuracy&Precision&recall&F1&Accuracy&Precision&recall&F1\\\hline
          4\_1&83.94&88.36&84.46&84.42&83.94&88.36&84.46&84.42\\\hline
          4\_2&73.40&79.14&78.76&77.23&73.40&79.14&78.76&77.23\\\hline
          4\_3&92.57&92.10&89.29&89.73&92.78&92.40&89.55&89.99\\\hline
          4\_4&84.98&87.90&82.81&82.74&84.98&87.90&82.81&82.74\\\hline
          5\_1&85.46&87.08&84.17&83.64&85.46&87.08&84.17&83.64\\\hline
          5\_2&82.61&91.05&67.85&68.47&82.77&91.15&67.95&68.58\\\hline
          5\_3&81.68&82.66&77.10&74.50&81.68&82.66&77.10&74.50\\\hline
          5\_4&82.37&85.15&81.98&82.44&82.37&85.15&81.98&82.44\\\hline
          5\_5&79.31&80.79&75.04&74.35&79.31&80.79&75.04&74.35\\\hline
          5\_6&91.19&93.19&91.95&92.15&93.52&94.23&93.04&93.24\\\hline
          Mean &83.75&86.74&81.34&80.97&84.02&86.89&81.49&81.11\\\hline
     \end{tabular}
      \caption{Results of team MedAIR for step recognition}
     \label{supMat:MedAIR_Step}
\end{table}

\subsubsection*{B.2 NUSControlLab}

\begin{table}[H]
    \centering
    \begin{tabular}{ | c | c |  c | c | c || c  | c | c | c |}
     \hline
          \multirow{2}{*}{Sequence}&\multicolumn{4}{c||}{Frame-by-Frame}&\multicolumn{4}{c|}{Application-Dependent}  \\\cline{2-9}
          &Accuracy&Precision&recall&F1&Accuracy&Precision&recall&F1\\\hline
          4\_1&96.22&95.17&94.69&94.65&96.41&95.40&94.94&94.92\\\hline
          4\_2&91.59&92.82&91.92&91.68&91.59&92.82&91.92&91.68\\\hline
          4\_3&98.87&98.31&98.21&98.23&99.12&98.59&98.54&98.54\\\hline
          4\_4&88.90&89.50&86.90&86.50&90.18&89.57&87.01&86.60\\\hline
          5\_1&98.41&98.59&97.97&98.15&98.62&98.96&98.33&98.51\\\hline
          5\_2&99.30&99.10&98.97&99.01&99.30&99.10&98.97&99.01\\\hline
          5\_3&89.18&86.98&78.80&79.44&89.18&86.98&78.80&79.44\\\hline
          5\_4&93.21&91.92&89.87&90.34&93.21&91.92&89.87&90.34\\\hline
          5\_5&95.44&94.59&94.01&93.94&95.78&94.99&94.46&94.41\\\hline
          5\_6&87.59&90.62&87.33&87.58&87.59&90.62&87.33&87.58\\\hline
          Mean &93.87&93.76&91.87&91.95&94.10&93.89&92.02&92.10\\\hline
     \end{tabular}
      \caption{Results of team NUSCONTROLLAB for phase recognition with multi granularity model}
     \label{supMat:NUSCONTROLLAB_Phase_Multi}
\end{table}

\begin{table}[H]
    \centering
    \begin{tabular}{ | c | c |  c | c | c || c  | c | c | c |}
     \hline
          \multirow{2}{*}{Sequence}&\multicolumn{4}{c||}{Frame-by-Frame}&\multicolumn{4}{c|}{Application-Dependent}  \\\cline{2-9}
          &Accuracy&Precision&recall&F1&Accuracy&Precision&recall&F1\\\hline
          4\_1&78.58&80.66&77.64&77.29&79.44&81.38&78.23&77.94\\\hline
          4\_2&87.49&92.22&84.22&86.24&88.07&92.54&84.53&86.54\\\hline
          4\_3&89.14&87.92&80.75&79.99&89.58&88.23&81.13&80.35\\\hline
          4\_4&69.77&76.66&51.30&56.92&71.20&78.05&52.13&57.72\\\hline
          5\_1&76.65&80.19&72.65&70.83&76.85&80.30&73.00&71.07\\\hline
          5\_2&63.83&46.05&47.62&45.44&63.83&46.05&47.62&45.44\\\hline
          5\_3&72.48&89.85&43.80&43.57&73.03&91.01&44.05&43.81\\\hline
          5\_4&89.35&91.37&86.41&87.02&91.02&92.39&87.39&87.92\\\hline
          5\_5&57.90&46.88&52.22&46.49&58.44&47.38&52.67&46.96\\\hline
          5\_6&54.98&63.99&49.79&50.86&54.98&63.99&49.79&50.86\\\hline
          Mean &74.02&75.58&64.64&64.46&74.64&76.13&65.05&64.86\\\hline
     \end{tabular}
      \caption{Results of team NUSCONTROLLAB for step\_multi recognition with multi granularity model}
     \label{supMat:NUSCONTROLLAB_Step_Multi}
\end{table}

\begin{table}[H]
    \centering
    \begin{tabular}{ | c | c |  c | c | c || c  | c | c | c |}
     \hline
          \multirow{2}{*}{Sequence}&\multicolumn{4}{c||}{Frame-by-Frame}&\multicolumn{4}{c|}{Application-Dependent}  \\\cline{2-9}
          &Accuracy&Precision&recall&F1&Accuracy&Precision&recall&F1\\\hline
          4\_1&58.02&64.93&67.40&63.55&60.57&67.42&69.83&65.95\\\hline
          4\_2&59.21&70.05&70.75&69.38&64.31&73.40&74.48&73.10\\\hline
          4\_3&59.93&70.83&71.79&69.06&62.72&73.37&74.44&71.90\\\hline
          4\_4&61.01&77.27&77.35&76.01&64.47&80.80&80.73&79.34\\\hline
          5\_1&57.90&75.53&78.27&76.60&58.95&76.62&79.42&77.73\\\hline
          5\_2&53.82&72.47&73.60&72.23&55.40&73.81&75.04&73.62\\\hline
          5\_3&53.37&71.94&72.89&71.17&56.00&74.36&74.98&73.43\\\hline
          5\_4&60.01&71.63&72.80&71.54&63.92&75.20&76.24&74.88\\\hline
          5\_5&60.81&74.33&76.29&73.83&65.56&77.75&79.88&77.53\\\hline
          5\_6&61.08&73.96&74.30&73.51&64.98&77.73&78.10&77.41\\\hline
          Mean &58.52&72.29&73.54&71.69&61.69&75.05&76.32&74.49\\\hline
     \end{tabular}
      \caption{Results of team NUSCONTROLLAB for activity recognition with multi-granularity model}
     \label{supMat:NUSCONTROLLAB_Activity_concat_Multi}
\end{table}

\begin{table}[H]
    \centering
    \begin{tabular}{ | c | c |  c | c | c || c  | c | c | c |}
     \hline
          \multirow{2}{*}{Sequence}&\multicolumn{4}{c||}{Frame-by-Frame}&\multicolumn{4}{c|}{Application-Dependent}  \\\cline{2-9}
          &Accuracy&Precision&recall&F1&Accuracy&Precision&recall&F1\\\hline
          4\_1&77.61&80.25&79.91&78.50&78.81&81.40&81.00&79.60\\\hline
          4\_2&79.43&85.03&82.30&82.43&81.32&86.26&83.64&83.77\\\hline
          4\_3&82.65&85.68&83.59&82.43&83.81&86.73&84.70&83.60\\\hline
          4\_4&73.23&81.14&71.85&73.14&75.28&82.81&73.29&74.55\\\hline
          5\_1&77.65&84.77&82.96&81.86&78.14&85.29&83.59&82.44\\\hline
          5\_2&72.31&72.54&73.40&72.22&72.84&72.99&73.88&72.69\\\hline
          5\_3&71.67&82.92&65.16&64.73&72.74&84.11&65.94&65.56\\\hline
          5\_4&80.86&84.97&83.03&82.96&82.72&86.50&84.50&84.38\\\hline
          5\_5&71.38&71.93&74.18&71.42&73.26&73.37&75.67&72.97\\\hline
          5\_6&67.88&76.19&70.47&70.65&69.18&77.45&71.74&71.95\\\hline
          Mean &75.47&80.54&76.68&76.03&76.81&81.69&77.80&77.15\\\hline
     \end{tabular}
      \caption{Results of team NUSCONTROLLAB for multi-granularity recognition}
     \label{supMat:NUSCONTROLLAB_Multi}
\end{table}

\subsubsection*{B.3 SK}

\begin{table}[H]
    \centering
    \begin{tabular}{ | c | c |  c | c | c || c  | c | c | c |}
     \hline
          \multirow{2}{*}{Sequence}&\multicolumn{4}{c||}{Frame-by-Frame}&\multicolumn{4}{c|}{Application-Dependent}  \\\cline{2-9}
          &Accuracy&Precision&recall&F1&Accuracy&Precision&recall&F1\\\hline
          4\_1&56.17&80.03&82.10&81.05&56.44&80.41&82.51&81.44\\\hline
          4\_2&54.35&80.65&77.19&77.32&54.41&80.71&77.31&77.43\\\hline
          4\_3&59.07&86.00&87.62&86.77&59.30&86.35&87.98&87.11\\\hline
          4\_4&62.00&88.15&90.56&89.27&62.19&88.38&90.81&89.52\\\hline
          5\_1&62.56&89.19&91.38&90.22&62.56&89.19&91.38&90.22\\\hline
          5\_2&56.96&84.00&84.53&84.16&57.03&84.10&84.66&84.29\\\hline
          5\_3&63.06&92.57&93.64&93.07&63.06&92.57&93.64&93.07\\\hline
          5\_4&55.80&78.09&80.71&79.29&55.80&78.09&80.71&79.29\\\hline
          5\_5&58.66&80.49&84.51&82.17&59.02&80.95&84.99&82.66\\\hline
          5\_6&59.85&78.74&83.35&80.64&60.09&79.03&83.68&80.96\\\hline
          Mean &58.85&83.79&85.56&84.40&58.99&83.98&85.77&84.60\\\hline
     \end{tabular}
      \caption{Results of team SK for phase recognition with multi-granularity model}
     \label{supMat:SK_Phase_Multi}
\end{table}

\begin{table}[H]
    \centering
    \begin{tabular}{ | c | c |  c | c | c || c  | c | c | c |}
     \hline
          \multirow{2}{*}{Sequence}&\multicolumn{4}{c||}{Frame-by-Frame}&\multicolumn{4}{c|}{Application-Dependent}  \\\cline{2-9}
          &Accuracy&Precision&recall&F1&Accuracy&Precision&recall&F1\\\hline
          4\_1&31.86&39.49&40.18&35.89&32.88&41.17&41.14&36.95\\\hline
          4\_2&35.24&60.74&55.44&56.35&36.03&61.27&55.95&56.85\\\hline
          4\_3&39.13&52.91&51.04&50.37&39.44&53.23&51.53&50.87\\\hline
          4\_4&41.76&50.16&53.29&49.41&42.02&50.37&53.51&49.62\\\hline
          5\_1&29.60&41.07&38.29&35.32&29.94&41.76&38.80&35.92\\\hline
          5\_2&32.06&60.35&44.38&46.51&32.52&60.72&44.64&46.69\\\hline
          5\_3&36.48&61.51&59.18&55.73&37.31&61.93&59.53&56.17\\\hline
          5\_4&34.94&51.11&39.44&34.96&35.06&51.36&39.67&35.31\\\hline
          5\_5&32.21&37.76&41.10&36.69&32.61&38.22&41.71&37.24\\\hline
          5\_6&40.72&51.76&44.13&41.05&40.72&51.76&44.13&41.05\\\hline
          Mean &35.40&50.69&46.65&44.23&35.85&51.18&47.06&44.67\\\hline
     \end{tabular}
      \caption{Results of team SK for step recognition with multi-granularity model}
     \label{supMat:SK_Step_Multi}
\end{table}

\begin{table}[H]
    \centering
    \begin{tabular}{ | c | c |  c | c | c || c  | c | c | c |}
     \hline
          \multirow{2}{*}{Sequence}&\multicolumn{4}{c||}{Frame-by-Frame}&\multicolumn{4}{c|}{Application-Dependent}  \\\cline{2-9}
          &Accuracy&Precision&recall&F1&Accuracy&Precision&recall&F1\\\hline
          4\_1&46.44&61.27&58.09&52.68&50.39&66.65&61.49&56.62\\\hline
          4\_2&50.09&66.19&65.42&61.98&55.28&69.81&69.05&65.84\\\hline
          4\_3&50.32&64.04&64.05&59.47&54.87&67.99&67.64&63.60\\\hline
          4\_4&48.63&68.86&64.98&61.29&54.52&73.13&69.77&66.64\\\hline
          5\_1&46.85&66.06&67.40&64.30&48.82&68.44&69.26&66.36\\\hline
          5\_2&43.12&65.07&67.08&62.39&44.83&67.33&68.58&64.06\\\hline
          5\_3&45.47&66.97&66.51&60.23&49.19&69.94&69.12&63.52\\\hline
          5\_4&53.16&66.33&67.35&62.99&56.08&69.72&69.98&66.02\\\hline
          5\_5&47.49&63.91&65.48&60.88&52.56&67.90&69.12&65.22\\\hline
          5\_6&53.14&63.77&63.49&60.19&57.40&67.90&67.65&64.71\\\hline
          Mean &48.47&65.25&64.99&60.64&52.40&68.88&68.17&64.26\\\hline
     \end{tabular}
      \caption{Results of team SK for activity recognition with multi-granularity model}
     \label{supMat:SK_Activity_concat_Multi}
\end{table}

\begin{table}[H]
    \centering
    \begin{tabular}{ | c | c |  c | c | c || c  | c | c | c |}
     \hline
          \multirow{2}{*}{Sequence}&\multicolumn{4}{c||}{Frame-by-Frame}&\multicolumn{4}{c|}{Application-Dependent}  \\\cline{2-9}
          &Accuracy&Precision&recall&F1&Accuracy&Precision&recall&F1\\\hline
          4\_1&44.82&60.27&60.13&56.54&46.57&62.75&61.72&58.33\\\hline
          4\_2&46.56&69.19&66.02&65.22&48.57&70.60&67.44&66.71\\\hline
          4\_3&49.50&67.65&67.57&65.54&51.21&69.19&69.05&67.19\\\hline
          4\_4&50.80&69.06&69.61&66.66&52.91&70.62&71.36&68.60\\\hline
          5\_1&46.34&65.44&65.69&63.28&47.11&66.46&66.48&64.16\\\hline
          5\_2&44.05&69.81&65.33&64.36&44.80&70.72&65.96&65.01\\\hline
          5\_3&48.34&73.68&73.11&69.67&49.85&74.82&74.09&70.92\\\hline
          5\_4&47.97&65.18&62.50&59.08&48.98&66.39&63.45&60.21\\\hline
          5\_5&46.12&60.72&63.70&59.92&48.06&62.35&65.27&61.71\\\hline
          5\_6&51.24&64.76&63.66&60.63&52.74&66.23&65.15&62.24\\\hline
          Mean &47.57&66.58&65.73&63.09&49.08&68.01&67.00&64.51\\\hline
     \end{tabular}
      \caption{Results of team SK for multi-granularity recognition}
     \label{supMat:SK_Multi}
\end{table}

\subsubsection*{B.4 UniandesBCV}
\begin{table}[H]
    \centering
    \begin{tabular}{ | c | c |  c | c | c || c  | c | c | c |}
     \hline
          \multirow{2}{*}{Sequence}&\multicolumn{4}{c||}{Frame-by-Frame}&\multicolumn{4}{c|}{Application-Dependent}  \\\cline{2-9}
          &Accuracy&Precision&recall&F1&Accuracy&Precision&recall&F1\\\hline
          4\_1&78.49&80.74&69.74&65.90&78.49&80.74&69.74&65.90\\\hline
          4\_2&81.92&83.19&63.01&63.15&82.10&83.28&63.37&63.56\\\hline
          4\_3&90.97&91.30&88.49&88.53&91.01&91.40&88.56&88.60\\\hline
          4\_4&87.12&85.76&79.49&78.88&87.12&85.76&79.49&78.88\\\hline
          5\_1&94.98&95.34&93.94&94.14&94.98&95.34&93.94&94.14\\\hline
          5\_2&95.28&95.66&94.46&94.66&95.39&95.78&94.59&94.79\\\hline
          5\_3&89.31&91.57&89.59&89.24&89.50&91.73&89.78&89.46\\\hline
          5\_4&89.92&90.04&87.45&87.49&90.53&91.22&88.37&88.43\\\hline
          5\_5&90.75&89.02&87.66&87.43&90.75&89.02&87.66&87.43\\\hline
          5\_6&94.38&95.76&93.71&94.14&94.67&96.35&94.21&94.64\\\hline
          Mean &89.31&89.84&84.75&84.36&89.45&90.06&84.97&84.58\\\hline
     \end{tabular}
      \caption{Results of team UniandesBCV for phase recognition}
     \label{supMat:UniandesBCV_Phase}
\end{table}

\begin{table}[H]
    \centering
    \begin{tabular}{ | c | c |  c | c | c || c  | c | c | c |}
     \hline
          \multirow{2}{*}{Sequence}&\multicolumn{4}{c||}{Frame-by-Frame}&\multicolumn{4}{c|}{Application-Dependent}  \\\cline{2-9}
          &Accuracy&Precision&recall&F1&Accuracy&Precision&recall&F1\\\hline
          4\_1&57.23&83.48&84.13&82.60&57.57&83.83&84.61&83.11\\\hline
          4\_2&56.97&84.65&85.82&85.22&56.97&84.65&85.82&85.22\\\hline
          4\_3&63.35&92.06&93.76&92.88&63.35&92.06&93.76&92.88\\\hline
          4\_4&59.86&86.50&87.95&86.66&59.86&86.50&87.95&86.66\\\hline
          5\_1&62.93&89.97&90.94&89.89&62.93&89.97&90.94&89.89\\\hline
          5\_2&62.37&91.50&92.17&91.71&62.37&91.50&92.17&91.71\\\hline
          5\_3&63.95&93.09&93.24&92.78&63.95&93.09&93.24&92.78\\\hline
          5\_4&59.05&82.62&84.86&83.41&59.05&82.62&84.86&83.41\\\hline
          5\_5&64.65&88.27&92.19&90.03&64.65&88.27&92.19&90.03\\\hline
          5\_6&63.33&83.27&87.37&84.58&63.76&83.74&88.05&85.21\\\hline
          Mean &61.37&87.54&89.24&87.98&61.45&87.62&89.36&88.09\\\hline
     \end{tabular}
      \caption{Results of team UniandesBCV for phase recognition with multi-granularity model}
     \label{supMat:UniandesBCV_Phase_Multi}
\end{table}

\begin{table}[H]
    \centering
    \begin{tabular}{ | c | c |  c | c | c || c  | c | c | c |}
     \hline
          \multirow{2}{*}{Sequence}&\multicolumn{4}{c||}{Frame-by-Frame}&\multicolumn{4}{c|}{Application-Dependent}  \\\cline{2-9}
          &Accuracy&Precision&recall&F1&Accuracy&Precision&recall&F1\\\hline
          4\_1&53.51&48.33&42.29&36.86&54.15&48.53&42.77&37.16\\\hline
          4\_2&47.82&60.60&44.62&46.94&47.82&60.60&44.62&46.94\\\hline
          4\_3&65.77&68.74&67.91&65.65&65.77&68.74&67.91&65.65\\\hline
          4\_4&65.34&74.67&65.99&66.07&65.34&74.67&65.99&66.07\\\hline
          5\_1&68.01&70.85&64.00&63.93&68.64&72.07&64.74&64.71\\\hline
          5\_2&57.81&78.08&58.97&60.98&58.32&78.59&59.38&61.32\\\hline
          5\_3&57.99&51.96&61.11&53.11&58.69&52.13&61.36&53.31\\\hline
          5\_4&60.93&64.04&55.21&47.25&60.93&64.04&55.21&47.25\\\hline
          5\_5&56.34&57.78&54.81&49.72&56.48&57.88&54.98&49.85\\\hline
          5\_6&65.28&75.69&56.92&55.27&65.94&75.93&57.34&55.53\\\hline
          Mean &59.88&65.07&57.18&54.58&60.21&65.32&57.43&54.78\\\hline
     \end{tabular}
      \caption{Results of team UniandesBCV for step recognition}
     \label{supMat:UniandesBCV_Step}
\end{table}

\begin{table}[H]
    \centering
    \begin{tabular}{ | c | c |  c | c | c || c  | c | c | c |}
     \hline
          \multirow{2}{*}{Sequence}&\multicolumn{4}{c||}{Frame-by-Frame}&\multicolumn{4}{c|}{Application-Dependent}  \\\cline{2-9}
          &Accuracy&Precision&recall&F1&Accuracy&Precision&recall&F1\\\hline
          4\_1&29.57&26.01&24.35&17.31&29.57&26.01&24.35&17.31\\\hline
          4\_2&38.44&49.47&48.02&46.95&38.65&49.60&48.14&47.03\\\hline
          4\_3&54.32&69.90&70.82&69.31&54.32&69.90&70.82&69.31\\\hline
          4\_4&44.08&55.93&48.26&50.99&45.19&56.91&49.17&51.83\\\hline
          5\_1&41.18&54.09&49.70&49.46&41.18&54.09&49.70&49.46\\\hline
          5\_2&30.99&49.44&24.91&26.64&31.01&49.51&24.94&26.69\\\hline
          5\_3&40.54&70.48&63.31&63.06&40.66&70.60&63.50&63.33\\\hline
          5\_4&42.09&53.95&49.63&51.00&42.91&54.63&50.32&51.64\\\hline
          5\_5&36.73&43.42&41.22&39.82&37.19&44.48&41.83&40.51\\\hline
          5\_6&38.22&51.01&51.97&51.09&38.41&51.22&52.18&51.30\\\hline
          Mean &39.62&52.37&47.22&46.56&39.91&52.69&47.49&46.84\\\hline
     \end{tabular}
      \caption{Results of team UniandesBCV for step recognition with multi-granularity model}
     \label{supMat:UniandesBCV_Step_Multi}
\end{table}

\begin{table}[H]
    \centering
    \begin{tabular}{ | c | c |  c | c | c || c  | c | c | c |}
     \hline
          \multirow{2}{*}{Sequence}&\multicolumn{4}{c||}{Frame-by-Frame}&\multicolumn{4}{c|}{Application-Dependent}  \\\cline{2-9}
          &Accuracy&Precision&recall&F1&Accuracy&Precision&recall&F1\\\hline
          4\_1&50.50&62.47&60.85&59.72&53.81&65.65&63.75&62.75\\\hline
          4\_2&51.94&68.28&64.45&65.07&57.60&73.13&69.39&70.07\\\hline
          4\_3&58.08&65.81&65.11&64.78&62.67&70.42&68.98&68.84\\\hline
          4\_4&57.09&69.32&67.20&67.52&64.87&75.02&72.82&73.13\\\hline
          5\_1&55.67&69.92&66.29&66.62&58.51&72.45&68.54&68.89\\\hline
          5\_2&58.95&72.92&69.41&70.17&63.84&76.34&72.94&73.62\\\hline
          5\_3&57.23&71.73&68.90&68.80&62.90&76.04&72.82&72.91\\\hline
          5\_4&53.05&64.67&60.30&59.76&59.18&69.54&64.81&64.61\\\hline
          5\_5&62.26&68.99&67.72&67.56&67.94&74.23&72.78&72.72\\\hline
          5\_6&52.64&64.48&60.51&60.57&61.79&71.35&67.02&67.38\\\hline
          Mean &55.74&67.86&65.08&65.06&61.31&72.42&69.39&69.49\\\hline
     \end{tabular}
      \caption{Results of team UniandesBCV for activity recognition}
     \label{supMat:UniandesBCV_Activity_concat}
\end{table}

\begin{table}[H]
    \centering
    \begin{tabular}{ | c | c |  c | c | c || c  | c | c | c |}
     \hline
          \multirow{2}{*}{Sequence}&\multicolumn{4}{c||}{Frame-by-Frame}&\multicolumn{4}{c|}{Application-Dependent}  \\\cline{2-9}
          &Accuracy&Precision&recall&F1&Accuracy&Precision&recall&F1\\\hline
          4\_1&54.60&62.20&63.02&60.74&56.37&64.05&64.80&62.48\\\hline
          4\_2&51.35&65.57&63.22&62.82&56.31&70.48&67.36&67.02\\\hline
          4\_3&57.43&66.44&66.64&65.60&59.81&68.51&68.60&67.55\\\hline
          4\_4&58.35&69.51&68.77&68.07&65.12&74.04&72.97&72.18\\\hline
          5\_1&57.25&70.50&69.78&68.87&59.34&72.20&71.32&70.46\\\hline
          5\_2&61.13&74.26&73.37&73.15&65.64&76.74&75.61&75.48\\\hline
          5\_3&57.52&72.45&72.45&71.27&61.10&75.23&74.92&73.85\\\hline
          5\_4&53.11&62.70&60.50&60.27&57.53&67.52&64.76&64.83\\\hline
          5\_5&60.56&68.54&67.35&66.98&66.53&72.91&71.42&71.07\\\hline
          5\_6&57.26&67.97&64.67&63.75&63.06&70.92&67.69&66.93\\\hline
          Mean &56.86&68.01&66.98&66.15&61.08&71.26&69.95&69.18\\\hline
     \end{tabular}
      \caption{Results of team UniandesBCV for activity recognition with multi-granularity model}
     \label{supMat:UniandesBCV_Activity_concat_Multi}
\end{table}

\begin{table}[H]
    \centering
    \begin{tabular}{ | c | c |  c | c | c || c  | c | c | c |}
     \hline
          \multirow{2}{*}{Sequence}&\multicolumn{4}{c||}{Frame-by-Frame}&\multicolumn{4}{c|}{Application-Dependent}  \\\cline{2-9}
          &Accuracy&Precision&recall&F1&Accuracy&Precision&recall&F1\\\hline
          4\_1&47.14&57.23&57.17&53.55&47.84&57.96&57.92&54.30\\\hline
          4\_2&48.92&66.57&65.69&65.00&50.64&68.24&67.10&66.43\\\hline
          4\_3&58.36&76.13&77.07&75.93&59.16&76.82&77.73&76.58\\\hline
          4\_4&54.10&70.65&68.33&68.57&56.72&72.48&70.03&70.22\\\hline
          5\_1&53.79&71.52&70.14&69.41&54.48&72.08&70.66&69.94\\\hline
          5\_2&51.50&71.74&63.48&63.83&53.01&72.58&64.24&64.63\\\hline
          5\_3&54.00&78.67&76.33&75.71&55.24&79.64&77.22&76.66\\\hline
          5\_4&51.42&66.42&64.99&64.89&53.16&68.26&66.64&66.63\\\hline
          5\_5&53.98&66.74&66.92&65.61&56.12&68.55&68.48&67.20\\\hline
          5\_6&52.94&67.42&68.01&66.47&55.08&68.62&69.31&67.81\\\hline
          Mean &52.61&69.31&67.81&66.90&54.15&70.53&68.93&68.04\\\hline
     \end{tabular}
      \caption{Results of team UniandesBCV for multi-granularity recognition}
     \label{supMat:UniandesBCV_Multi}
\end{table}

\subsubsection*{B.5 Wr0112358}

\begin{table}[H]
    \centering
    \begin{tabular}{ | c | c |  c | c | c || c  | c | c | c |}
     \hline
          \multirow{2}{*}{Sequence}&\multicolumn{4}{c||}{Frame-by-Frame}&\multicolumn{4}{c|}{Application-Dependent}  \\\cline{2-9}
          &Accuracy&Precision&recall&F1&Accuracy&Precision&recall&F1\\\hline
          4\_1&96.70&96.08&95.94&95.93&97.00&96.52&96.38&96.38\\\hline
          4\_2&88.69&85.29&78.97&79.69&88.69&85.29&78.97&79.69\\\hline
          4\_3&93.13&92.39&91.25&91.08&93.13&92.39&91.25&91.08\\\hline
          4\_4&80.57&89.94&89.80&89.72&80.77&90.22&90.09&90.00\\\hline
          5\_1&99.02&98.58&98.51&98.53&99.31&98.89&98.87&98.87\\\hline
          5\_2&94.97&94.68&94.10&94.02&95.08&94.80&94.23&94.15\\\hline
          5\_3&94.22&96.86&96.65&96.69&94.22&96.86&96.65&96.69\\\hline
          5\_4&93.54&91.88&91.08&91.28&93.87&92.32&91.48&91.70\\\hline
          5\_5&87.60&85.84&83.50&82.93&88.12&86.37&84.18&83.70\\\hline
          5\_6&85.38&94.29&94.17&93.93&85.83&94.78&94.67&94.44\\\hline
          Mean &91.38&92.58&91.39&91.38&91.60&92.85&91.68&91.67\\\hline
     \end{tabular}
      \caption{Results of team wr0112358 for phase recognition}
     \label{supMat:wr0112358_Phase}
\end{table}

\begin{table}[H]
    \centering
    \begin{tabular}{ | c | c |  c | c | c || c  | c | c | c |}
     \hline
          \multirow{2}{*}{Sequence}&\multicolumn{4}{c||}{Frame-by-Frame}&\multicolumn{4}{c|}{Application-Dependent}  \\\cline{2-9}
          &Accuracy&Precision&recall&F1&Accuracy&Precision&recall&F1\\\hline
          4\_1&92.02&93.30&92.77&92.77&92.49&93.94&93.47&93.47\\\hline
          4\_2&87.22&86.33&85.79&85.54&87.22&86.33&85.79&85.54\\\hline
          4\_3&92.55&89.82&88.26&88.42&92.55&89.82&88.26&88.42\\\hline
          4\_4&79.75&90.87&90.01&89.84&79.97&91.34&90.38&90.19\\\hline
          5\_1&80.68&81.36&67.18&65.60&80.79&81.47&67.31&65.71\\\hline
          5\_2&82.18&82.42&68.26&67.40&82.18&82.42&68.26&67.40\\\hline
          5\_3&86.15&88.25&82.26&82.91&86.33&88.34&82.44&83.03\\\hline
          5\_4&88.28&86.37&81.00&81.05&88.28&86.37&81.00&81.05\\\hline
          5\_5&80.45&82.80&72.17&70.41&81.29&82.85&72.29&70.51\\\hline
          5\_6&73.54&82.42&70.05&69.86&73.84&82.57&70.55&70.42\\\hline
          Mean &84.28&86.39&79.77&79.38&84.49&86.55&79.97&79.58\\\hline
     \end{tabular}
      \caption{Results of team wr0112358 for phase recognition with multi-granularity model}
     \label{supMat:wr0112358_Phase_Multi}
\end{table}

\begin{table}[H]
    \centering
    \begin{tabular}{ | c | c |  c | c | c || c  | c | c | c |}
     \hline
          \multirow{2}{*}{Sequence}&\multicolumn{4}{c||}{Frame-by-Frame}&\multicolumn{4}{c|}{Application-Dependent}  \\\cline{2-9}
          &Accuracy&Precision&recall&F1&Accuracy&Precision&recall&F1\\\hline
          4\_1&48.31&60.44&44.32&39.87&48.74&60.67&44.58&40.31\\\hline
          4\_2&53.56&70.44&58.83&60.73&53.83&71.00&59.47&61.44\\\hline
          4\_3&74.05&74.11&66.28&66.19&75.26&75.29&67.58&67.55\\\hline
          4\_4&63.96&75.81&73.81&71.93&64.22&76.04&74.02&72.13\\\hline
          5\_1&63.12&71.99&66.20&63.11&64.18&72.57&66.92&64.15\\\hline
          5\_2&55.35&72.80&48.00&51.24&55.55&72.96&48.18&51.42\\\hline
          5\_3&66.02&77.88&75.03&72.36&66.61&78.07&75.33&72.75\\\hline
          5\_4&68.26&74.80&62.87&61.20&70.12&77.14&64.48&62.94\\\hline
          5\_5&67.59&69.39&67.68&67.29&69.44&71.08&69.38&69.14\\\hline
          5\_6&69.13&75.91&74.20&72.50&69.46&76.24&74.58&72.88\\\hline
          Mean &62.93&72.36&63.72&62.64&63.74&73.11&64.45&63.47\\\hline
     \end{tabular}
      \caption{Results of team wr0112358 for step recognition}
     \label{supMat:wr0112358_Step}
\end{table}

\begin{table}[H]
    \centering
    \begin{tabular}{ | c | c |  c | c | c || c  | c | c | c |}
     \hline
          \multirow{2}{*}{Sequence}&\multicolumn{4}{c||}{Frame-by-Frame}&\multicolumn{4}{c|}{Application-Dependent}  \\\cline{2-9}
          &Accuracy&Precision&recall&F1&Accuracy&Precision&recall&F1\\\hline
          4\_1&52.21&53.95&53.06&50.55&53.24&55.07&54.32&51.74\\\hline
          4\_2&54.59&63.70&53.70&53.39&55.50&64.55&54.51&54.20\\\hline
          4\_3&60.68&58.14&49.27&48.19&60.94&58.32&49.62&48.59\\\hline
          4\_4&47.58&43.71&47.21&42.09&50.91&44.78&48.52&43.18\\\hline
          5\_1&48.80&32.67&37.34&29.55&49.40&34.75&38.11&30.64\\\hline
          5\_2&46.35&51.18&36.89&29.77&46.76&51.33&37.11&29.94\\\hline
          5\_3&54.31&64.53&56.11&57.04&54.50&64.63&56.28&57.21\\\hline
          5\_4&56.16&49.91&47.55&43.04&57.97&51.24&48.76&44.28\\\hline
          5\_5&44.17&45.69&41.14&36.42&44.53&45.75&41.26&36.51\\\hline
          5\_6&38.39&36.22&35.70&32.18&40.37&37.59&36.91&33.39\\\hline
          Mean &50.32&49.97&45.80&42.22&51.41&50.80&46.54&42.97\\\hline
     \end{tabular}
      \caption{Results of team wr0112358 for step recognition with multi-granularity model}
     \label{supMat:wr0112358_Step_Multi}
\end{table}

\begin{table}[H]
    \centering
    \begin{tabular}{ | c | c |  c | c | c || c  | c | c | c |}
     \hline
          \multirow{2}{*}{Sequence}&\multicolumn{4}{c||}{Frame-by-Frame}&\multicolumn{4}{c|}{Application-Dependent}  \\\cline{2-9}
          &Accuracy&Precision&recall&F1&Accuracy&Precision&recall&F1\\\hline
          4\_1&54.33&64.72&61.89&58.69&57.55&67.86&64.23&61.38\\\hline
          4\_2&45.76&60.17&57.61&54.57&50.98&64.38&61.32&58.58\\\hline
          4\_3&57.34&66.99&68.38&65.72&61.73&71.67&72.02&69.50\\\hline
          4\_4&55.34&71.18&69.68&66.83&61.26&75.84&74.25&71.74\\\hline
          5\_1&51.78&68.94&65.28&64.34&53.93&70.81&66.83&65.99\\\hline
          5\_2&52.43&70.06&70.57&68.71&55.06&72.44&72.50&70.87\\\hline
          5\_3&54.57&69.93&70.44&67.97&58.59&73.28&73.41&71.18\\\hline
          5\_4&57.23&66.64&66.80&65.16&60.91&70.14&70.03&68.55\\\hline
          5\_5&61.67&72.55&73.28&71.74&66.59&76.59&77.16&75.86\\\hline
          5\_6&58.41&71.08&69.31&67.92&62.94&74.76&73.09&71.97\\\hline
          Mean &54.89&68.22&67.32&65.17&58.95&71.78&70.49&68.56\\\hline
     \end{tabular}
      \caption{Results of team wr0112358 for activity recognition}
     \label{supMat:wr0112358_Activity_concat}
\end{table}

\begin{table}[H]
    \centering
    \begin{tabular}{ | c | c |  c | c | c || c  | c | c | c |}
     \hline
          \multirow{2}{*}{Sequence}&\multicolumn{4}{c||}{Frame-by-Frame}&\multicolumn{4}{c|}{Application-Dependent}  \\\cline{2-9}
          &Accuracy&Precision&recall&F1&Accuracy&Precision&recall&F1\\\hline
          4\_1&54.94&66.90&61.73&59.66&57.94&69.46&64.23&62.44\\\hline
          4\_2&57.88&69.85&66.74&66.24&61.89&74.31&70.78&70.39\\\hline
          4\_3&59.19&71.73&67.07&66.89&61.99&74.51&69.73&69.65\\\hline
          4\_4&54.62&73.29&69.04&68.33&59.69&77.27&73.36&72.81\\\hline
          5\_1&46.30&69.16&59.19&58.51&47.50&70.27&60.22&59.62\\\hline
          5\_2&50.72&71.25&64.76&64.56&52.43&73.18&66.30&66.27\\\hline
          5\_3&50.74&72.68&66.99&66.59&53.12&75.06&69.04&68.79\\\hline
          5\_4&54.11&67.54&63.77&62.78&56.39&69.37&65.56&64.65\\\hline
          5\_5&54.17&72.56&65.42&64.63&57.10&75.11&68.03&67.45\\\hline
          5\_6&56.18&73.24&66.64&66.44&59.05&76.18&69.64&69.55\\\hline
          Mean &53.88&70.82&65.13&64.46&56.71&73.47&67.69&67.16\\\hline
     \end{tabular}
      \caption{Results of team wr0112358 for activity recognition with multi-granularity model}
     \label{supMat:wr0112358_Activity_concat_Multi}
\end{table}

\begin{table}[H]
    \centering
    \begin{tabular}{ | c | c |  c | c | c || c  | c | c | c |}
     \hline
          \multirow{2}{*}{Sequence}&\multicolumn{4}{c||}{Frame-by-Frame}&\multicolumn{4}{c|}{Application-Dependent}  \\\cline{2-9}
          &Accuracy&Precision&recall&F1&Accuracy&Precision&recall&F1\\\hline
          4\_1&66.39&71.38&69.19&67.66&67.89&72.82&70.67&69.22\\\hline
          4\_2&66.56&73.29&68.74&68.39&68.20&75.06&70.36&70.05\\\hline
          4\_3&70.81&73.23&68.20&67.84&71.83&74.22&69.20&68.89\\\hline
          4\_4&60.65&69.29&68.76&66.75&63.53&71.13&70.75&68.73\\\hline
          5\_1&58.59&61.06&54.57&51.22&59.23&62.16&55.21&51.99\\\hline
          5\_2&59.75&68.28&56.64&53.91&60.45&68.98&57.22&54.54\\\hline
          5\_3&63.74&75.15&68.45&68.85&64.65&76.01&69.25&69.68\\\hline
          5\_4&66.19&67.94&64.11&62.29&67.55&69.00&65.11&63.33\\\hline
          5\_5&59.59&67.02&59.58&57.15&60.97&67.90&60.53&58.15\\\hline
          5\_6&56.04&63.96&57.46&56.16&57.75&65.45&59.03&57.79\\\hline
          Mean &62.83&69.06&63.57&62.02&64.21&70.27&64.73&63.24\\\hline
     \end{tabular}
      \caption{Results of team wr0112358 for multi-granularity recognition}
     \label{supMat:wr0112358_Multi}
\end{table}

\subsubsection*{B.6 IMPACT}
\begin{table}[H]
    \centering
    \begin{tabular}{ | c | c |  c | c | c || c  | c | c | c |}
     \hline
          \multirow{2}{*}{Sequence}&\multicolumn{4}{c||}{Frame-by-Frame}&\multicolumn{4}{c|}{Application-Dependent}  \\\cline{2-9}
          &Accuracy&Precision&recall&F1&Accuracy&Precision&recall&F1\\\hline
          4\_1&78.02&76.63&76.38&76.43&84.21&76.97&76.86&76.87\\\hline
          4\_2&81.86&78.32&74.46&75.34&81.86&78.32&74.46&75.34\\\hline
          4\_3&82.71&76.16&74.80&75.17&82.98&76.57&75.15&75.53\\\hline
          4\_4&79.45&82.34&81.98&81.80&79.69&82.63&82.31&82.14\\\hline
          5\_1&82.24&77.38&74.78&75.02&82.24&77.38&74.78&75.02\\\hline
          5\_2&87.50&83.29&81.32&81.94&87.50&83.29&81.32&81.94\\\hline
          5\_3&81.49&79.12&72.39&73.18&81.59&79.19&72.48&73.26\\\hline
          5\_4&79.87&73.58&70.12&70.50&79.87&73.58&70.12&70.50\\\hline
          5\_5&78.56&69.47&69.46&69.34&78.87&69.83&69.86&69.76\\\hline
          5\_6&67.77&82.87&83.22&82.69&67.77&82.87&83.22&82.69\\\hline
          Mean &79.95&77.92&75.89&76.14&80.66&78.06&76.06&76.30\\\hline
     \end{tabular}
      \caption{Results of team Impact for phase recognition}
     \label{supMat:Impact_Phase}
\end{table}

\begin{table}[H]
    \centering
    \begin{tabular}{ | c | c |  c | c | c || c  | c | c | c |}
     \hline
          \multirow{2}{*}{Sequence}&\multicolumn{4}{c||}{Frame-by-Frame}&\multicolumn{4}{c|}{Application-Dependent}  \\\cline{2-9}
          &Accuracy&Precision&recall&F1&Accuracy&Precision&recall&F1\\\hline
          4\_1&84.46&83.64&76.49&75.41&84.62&83.89&76.72&75.63\\\hline
          4\_2&85.85&82.48&79.75&80.64&85.85&82.48&79.75&80.64\\\hline
          4\_3&77.74&72.48&65.84&66.61&77.74&72.48&65.84&66.61\\\hline
          4\_4&86.63&81.07&80.61&80.68&86.63&81.07&80.61&80.68\\\hline
          5\_1&81.02&73.82&72.34&72.63&81.17&74.02&72.59&72.88\\\hline
          5\_2&85.53&82.07&76.84&77.97&85.53&82.07&76.84&77.97\\\hline
          5\_3&86.20&85.26&73.77&74.77&86.34&85.31&73.91&74.85\\\hline
          5\_4&86.24&83.51&79.45&80.22&86.24&83.51&79.45&80.22\\\hline
          5\_5&81.96&74.82&73.99&73.97&81.96&74.82&73.99&73.97\\\hline
          5\_6&70.96&84.87&79.45&79.41&70.96&84.87&79.45&79.41\\\hline
          Mean &82.66&80.40&75.85&76.23&82.70&80.45&75.91&76.29\\\hline
     \end{tabular}
      \caption{Results of team Impact for phase recognition with multi-granularity model}
     \label{supMat:Impact_Phase_Multi}
\end{table}

\begin{table}[H]
    \centering
    \begin{tabular}{ | c | c |  c | c | c || c  | c | c | c |}
     \hline
          \multirow{2}{*}{Sequence}&\multicolumn{4}{c||}{Frame-by-Frame}&\multicolumn{4}{c|}{Application-Dependent}  \\\cline{2-9}
          &Accuracy&Precision&recall&F1&Accuracy&Precision&recall&F1\\\hline
          4\_1&47.94&37.63&42.29&35.63&49.10&38.99&43.58&36.76\\\hline
          4\_2&43.53&58.02&47.84&51.10&44.26&58.45&48.44&51.60\\\hline
          4\_3&43.27&42.81&40.94&41.04&44.77&44.48&42.09&42.38\\\hline
          4\_4&43.37&53.27&49.57&51.02&44.75&53.99&50.40&51.78\\\hline
          5\_1&42.66&36.93&37.77&34.50&42.95&37.39&38.08&34.88\\\hline
          5\_2&39.80&35.89&34.85&31.38&39.80&35.89&34.85&31.38\\\hline
          5\_3&55.65&62.13&58.09&57.21&55.91&62.29&58.29&57.41\\\hline
          5\_4&49.36&36.77&38.74&32.99&50.79&37.25&39.49&33.60\\\hline
          5\_5&30.38&36.20&32.61&33.32&30.86&36.73&33.29&33.95\\\hline
          5\_6&59.98&69.06&58.68&59.25&61.63&69.75&59.56&60.29\\\hline
          Mean &45.59&46.87&44.14&42.74&46.48&47.52&44.81&43.40\\\hline
     \end{tabular}
      \caption{Results of team Impact for step recognition}
     \label{supMat:Impact_Step}
\end{table}

\begin{table}[H]
    \centering
    \begin{tabular}{ | c | c |  c | c | c || c  | c | c | c |}
     \hline
          \multirow{2}{*}{Sequence}&\multicolumn{4}{c||}{Frame-by-Frame}&\multicolumn{4}{c|}{Application-Dependent}  \\\cline{2-9}
          &Accuracy&Precision&recall&F1&Accuracy&Precision&recall&F1\\\hline
          4\_1&60.98&60.31&57.90&52.59&61.66&60.81&58.49&53.13\\\hline
          4\_2&55.42&64.61&56.28&55.73&55.42&64.61&56.28&55.73\\\hline
          4\_3&56.94&48.83&51.18&47.66&57.92&49.08&51.51&47.89\\\hline
          4\_4&61.38&55.41&56.51&54.90&62.00&56.20&57.02&55.51\\\hline
          5\_1&54.05&47.13&42.11&40.15&54.20&47.63&42.37&40.60\\\hline
          5\_2&53.49&53.13&40.59&40.10&53.49&53.13&40.59&40.10\\\hline
          5\_3&59.52&65.67&59.15&53.64&59.55&65.68&59.18&53.65\\\hline
          5\_4&59.11&56.39&47.78&43.51&60.63&57.67&48.93&44.49\\\hline
          5\_5&52.02&51.61&51.13&49.52&52.02&51.61&51.13&49.52\\\hline
          5\_6&53.85&58.92&42.87&38.06&53.93&58.98&42.95&38.12\\\hline
          Mean &56.68&56.20&50.55&47.59&57.08&56.54&50.84&47.88\\\hline
     \end{tabular}
      \caption{Results of team Impact for step recognition with multi-granularity model}
     \label{supMat:Impact_Step_Multi}
\end{table}

\begin{table}[H]
    \centering
    \begin{tabular}{ | c | c |  c | c | c || c  | c | c | c |}
     \hline
          \multirow{2}{*}{Sequence}&\multicolumn{4}{c||}{Frame-by-Frame}&\multicolumn{4}{c|}{Application-Dependent}  \\\cline{2-9}
          &Accuracy&Precision&recall&F1&Accuracy&Precision&recall&F1\\\hline
          4\_1&56.87&65.08&57.10&56.31&59.78&68.68&59.46&58.68\\\hline
          4\_2&56.83&69.91&64.42&62.96&59.74&72.91&67.17&65.65\\\hline
          4\_3&59.56&68.94&61.80&61.44&62.26&71.80&64.36&64.15\\\hline
          4\_4&37.65&57.26&43.51&46.65&39.94&59.44&45.51&48.55\\\hline
          5\_1&52.17&68.69&59.86&61.03&53.83&70.07&61.00&62.17\\\hline
          5\_2&54.06&70.00&63.08&63.90&56.63&71.76&64.85&65.69\\\hline
          5\_3&55.73&74.87&67.40&67.20&58.23&77.38&69.40&69.21\\\hline
          5\_4&59.03&72.08&65.94&65.07&61.66&74.42&68.11&67.29\\\hline
          5\_5&57.41&69.71&60.88&61.52&60.89&72.37&63.34&64.20\\\hline
          5\_6&64.95&78.86&70.92&71.41&68.02&81.92&73.54&74.18\\\hline
          Mean &55.43&69.54&61.49&61.75&58.10&72.07&63.67&63.98\\\hline
     \end{tabular}
      \caption{Results of team Impact for activity recognition}
     \label{supMat:Impact_Activity_concat}
\end{table}

\begin{table}[H]
    \centering
    \begin{tabular}{ | c | c |  c | c | c || c  | c | c | c |}
     \hline
          \multirow{2}{*}{Sequence}&\multicolumn{4}{c||}{Frame-by-Frame}&\multicolumn{4}{c|}{Application-Dependent}  \\\cline{2-9}
          &Accuracy&Precision&recall&F1&Accuracy&Precision&recall&F1\\\hline
          4\_1&51.14&62.23&53.31&53.86&54.15&65.26&55.77&56.44\\\hline
          4\_2&55.56&66.52&60.57&60.03&58.53&68.98&63.07&62.50\\\hline
          4\_3&59.97&72.29&62.90&63.27&63.13&75.66&65.67&66.18\\\hline
          4\_4&51.57&68.93&58.94&61.28&54.67&71.94&61.95&64.28\\\hline
          5\_1&57.89&73.59&62.16&64.80&59.55&75.04&63.63&66.24\\\hline
          5\_2&59.20&73.92&63.21&65.74&62.24&75.89&65.00&67.58\\\hline
          5\_3&59.61&76.11&62.89&63.76&62.04&77.99&64.53&65.44\\\hline
          5\_4&62.36&74.01&68.27&68.06&65.49&76.36&70.57&70.43\\\hline
          5\_5&57.24&66.90&59.30&60.63&60.88&70.07&62.25&63.55\\\hline
          5\_6&66.84&78.96&68.35&70.43&69.88&81.85&71.03&73.25\\\hline
          Mean &58.14&71.35&61.99&63.19&61.06&73.90&64.35&65.59\\\hline
     \end{tabular}
      \caption{Results of team Impact for activity recognition with multi-granularity model}
     \label{supMat:Impact_Activity_concat_Multi}
\end{table}

\begin{table}[H]
    \centering
    \begin{tabular}{ | c | c |  c | c | c || c  | c | c | c |}
     \hline
          \multirow{2}{*}{Sequence}&\multicolumn{4}{c||}{Frame-by-Frame}&\multicolumn{4}{c|}{Application-Dependent}  \\\cline{2-9}
          &Accuracy&Precision&recall&F1&Accuracy&Precision&recall&F1\\\hline
          4\_1&65.53&68.73&62.57&60.62&66.81&69.99&63.66&61.73\\\hline
          4\_2&65.61&71.20&65.53&65.47&66.60&72.02&66.37&66.29\\\hline
          4\_3&64.88&64.53&59.97&59.18&66.26&65.74&61.01&60.22\\\hline
          4\_4&66.53&68.47&65.35&65.62&67.77&69.73&66.53&66.82\\\hline
          5\_1&64.32&64.85&58.87&59.19&64.97&65.56&59.53&59.91\\\hline
          5\_2&66.07&69.71&60.21&61.27&67.08&70.36&60.81&61.88\\\hline
          5\_3&68.44&75.68&65.27&64.05&69.31&76.33&65.87&64.65\\\hline
          5\_4&69.24&71.30&65.17&63.93&70.78&72.51&66.32&65.05\\\hline
          5\_5&63.74&64.44&61.47&61.38&64.95&65.50&62.46&62.35\\\hline
          5\_6&63.88&74.25&63.56&62.63&64.92&75.23&64.48&63.60\\\hline
          Mean &65.82&69.32&62.80&62.33&66.95&70.30&63.70&63.25\\\hline
     \end{tabular}
      \caption{Results of team Impact for multi-granularity recognition}
     \label{supMat:Impact_Multi}
\end{table}

\subsection*{C. MISAW data annotation protocol}
The annotation protocol given to each observer is the following.


\section*{Supplementary material}

# MISAW data annotation protocol

Each session of MISAW (MIcro-Surgical Anastomose Workflow recognition on training sessions) dataset must been manually annotated thanks to *Surgery Workflow Toolbox* [Annotate] software to generate procedural sequences. The sequence have to been annotate at the three following levels of granularity [1]:

- Phases: the main periods of the intervention. A phase is composed of one or more steps.
- Steps: a step is a sequence of activities used to achieve a surgical objective
- Activities: An activity is a physical action performed by the surgeon and is composed of three components:
  - Action verb: describe the gesture.
  - Target: what is affected by the action.
  - Surgical instrument: with what the action is realized.

## I.    Phases & steps

The task is composed of 2 phases (in bold), each of them composed of 3 steps (in italic):

- **Suturing**: joining artificial vessels by using sterile suture material and a needle
  - *Needle holding*: Put the needle on place to allow the suturing
  - *Suture making*: Insert and pull the needle into artificial vessels.
  - *Suture handling*: Put the artificial vessels align and the wire to allow the Knot Tying
- **Knot Tying**: make 3 knots to keep the 2 artificial vessels close.
  - *1° Knot*: First knot
  - *2° Knot*: Second knot
  - *3° Knot*: Third knot

| Phases | Steps | Start | End | Illustration |
|--------|-------|-------|-----|--------------|
| Suturing | *Needle holding* | First activity | When the needle is place to start the insertion |  |
| | *Suture making* | Insert the needle on one vessel | The needle completely pass through both vessels |  |

| | | | | |
|---|---|---|---|---|
| | *Suture handling* | The needle completely pass through both vessels | Stop pulling the wire through vessels | 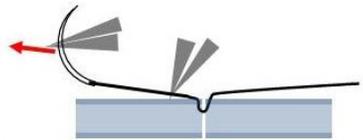 |
| **Knot Tying** | *1° knot* | Stop pulling the wire through vessels | Stop pulling the 2 part of the wire to tie the 1° knot | 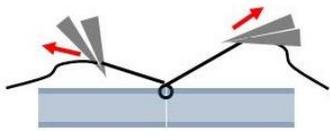 |
| | *2° knot* | Stop pulling the 2 part of the wire to tie the 1° knot | Stop pulling the 2 part of the wire to tie the 2° knot | 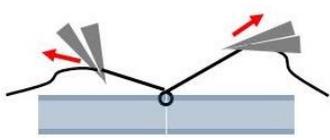 |
| | *3° knot* | Stop pulling the 2 part of the wire to tie the 2° knot | End of video | 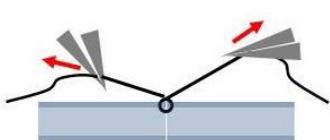 |

## II.    Activities

Activities where defined thanks to 17 action verbs, 9 targets, and 1 surgical instrument:

### A.    Action Verbs:

- Catch: when tooltips begin to close to grab a target between them.
- Give slack: move a target (a wire or a part of it) to have more space to work
- Hold: keep a target between tooltips (it is a transition action verb between other verbs) or block a vessel by applying a force from the target interior or exterior.
- Insert: Pierce a membrane of a vessel with a needle
- Loosen completely: undo completely a knot
- Loosen partially: undo partial a knot (generally to tighten it up without making again wire loop)
- Make a loop: action consisting to create a loop with a wire around the second tool. Both tooltips exert it at the same time.
- Pass through: action consisting to move a tool through a wire loop.
- Position: replace the needle held by a tooltip with the other tooltip.
- Pull: to exert force on a target to draw it towards the source of the force

See bellow examples of holding vessels. Black do represent tool tips, green circle a vessel, black line the background.

| | Cases of holding a vessel | Illustration |
|---|---|---|
| **Hold** | applying force from the target interior by opening tool | 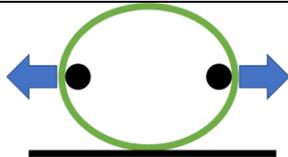 |
| | applying force from the target exterior by closing tool | 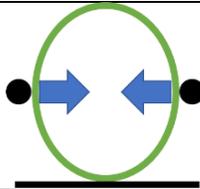 |
| | applying force by pressing between tool and background. | 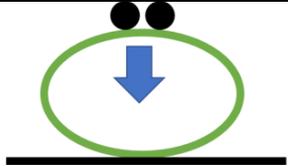 |

### B.     Targets:

- Needle
- Wire (defined until end of suturing phase)
- Both artificial vessels: left and right artificial vessels are the target at the same time of the action verb
- Left artificial vessel
- Right artificial vessel
- Long wire strand (defined from the beginning of first knot)
- Short wire strand (defined from the beginning of first knot)
- Wire loop: a loop creates to make the knot (only use with action verb "pass through")
- Knot

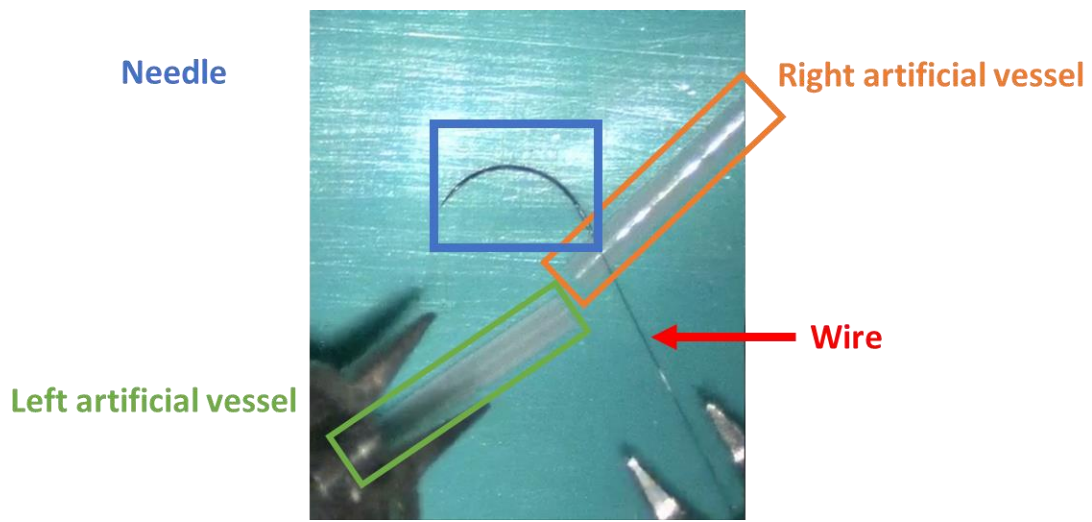

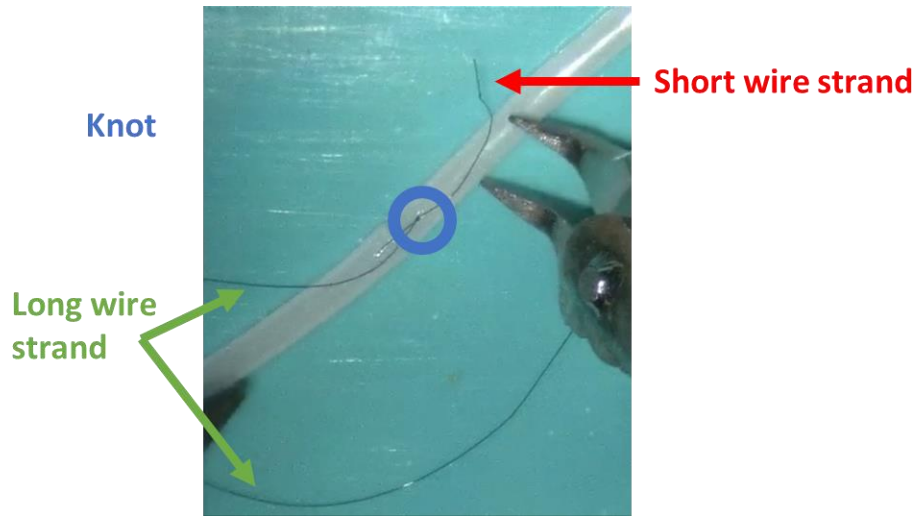

**Knot**

**Short wire strand**

**Long wire strand**

    Surgical instrument

Surgical instrument is always the same. No differences were done between left and right surgical instrument. Both are needle holders



\end{document}